\pdfoutput=1

\documentclass[10pt,twocolumn,letterpaper]{article}

\newcommand{\paperrepourl}{https://github.com/unclearness/cross3dgs}

\newcommand{\paperauthors}{%
  Hiroyuki Sato\\
  MinD in a Device Co., Ltd., Tokyo, Japan\\
  {\tt\small h.sato@mindinadevice.com}
  \and
  Yusuke Takimoto\\
  Independent Researcher, Kanagawa, Japan\\
  {\tt\small takiyu1025.txu@gmail.com}%
}

  \usepackage[pagenumbers]{wacv}     %

\usepackage{array}
\usepackage{algorithm}
\usepackage{algpseudocode}
\usepackage{xcolor}
\usepackage{multirow}

\definecolor{gpuNV}{HTML}{4F8000}
\definecolor{gpuIntel}{HTML}{0083B0}
\definecolor{gpuArm}{HTML}{C76A00}
\definecolor{gpuQC}{HTML}{3253DC}
\definecolor{gpuAMD}{HTML}{ED1C24}
\definecolor{gpuM5}{HTML}{8F6E00}
\definecolor{gpuA14}{HTML}{963D97}
\definecolor{gpuIntelUHD}{HTML}{005A87}
\newcommand{\gPRO}{\textcolor{gpuNV}{PRO6000}}
\newcommand{\gIris}{\textcolor{gpuIntel}{Iris Xe}}
\newcommand{\gGone}{\textcolor{gpuArm}{G1-Ultra}}
\newcommand{\gAdreno}{\textcolor{gpuQC}{Adreno740}}
\newcommand{\gRX}{\textcolor{gpuAMD}{RX9070XT}}
\newcommand{\gMfive}{\textcolor{gpuM5}{M5Pro}}
\newcommand{\gAfourteen}{\textcolor{gpuA14}{A14}}
\newcommand{\gUHD}{\textcolor{gpuIntelUHD}{UHD730}}

\newcommand{\twol}[2]{\begin{tabular}[t]{@{}c@{}}#1\\#2\end{tabular}}

\usepackage{float}
\usepackage{placeins}
\newcommand{\supplements}[1]{\noindent\textit{Supplements, in the main paper: #1.}\par\smallskip}

\definecolor{wacvblue}{rgb}{0.21,0.49,0.74}
\usepackage[pagebackref,breaklinks,colorlinks,allcolors=wacvblue]{hyperref}

\def\confName{WACV}
\def\confYear{2027}

\title{ Cross3DGS: Unified Vulkan-Based 3D Gaussian Splatting Training from Workstations to Mobile Devices}

\author{\paperauthors}

\begin{document}
\twocolumn[{%
  \renewcommand\twocolumn[1][]{#1}%
  \maketitle
  \begingroup
    \centering
    \includegraphics[width=\textwidth]{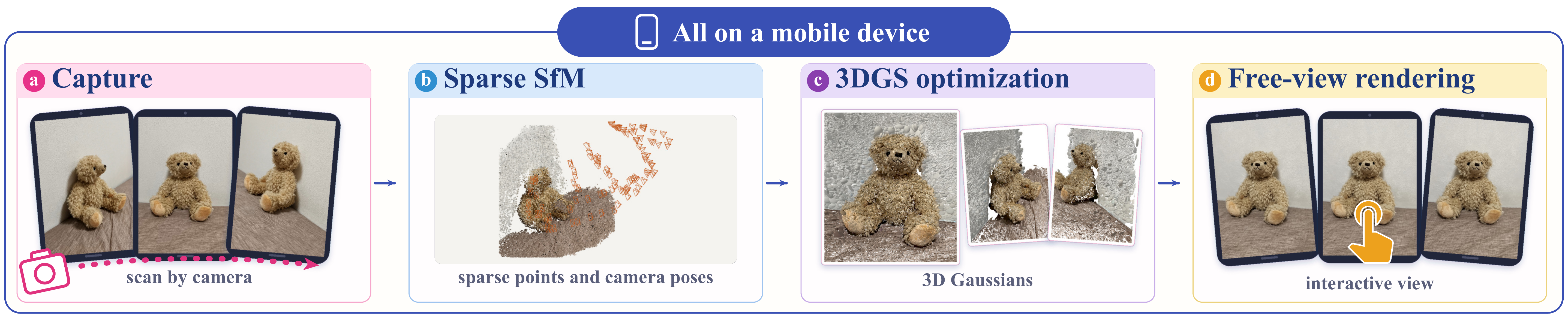}
    \captionsetup{hypcap=false}
    \captionof{figure}{An end-to-end 3DGS reconstruction application for mobile devices, shown on a smartphone: capture, sparse structure from motion, on-device 3DGS
    optimization, and interactive viewing. The application is available for both Android and iOS. Section~\ref{sec:eval:mobile} describes the details of this application.}
    \label{fig:teaser}
    \par
  \endgroup
  \vspace{4pt}%
}]

\begin{abstract}
Scenes captured with handheld and wearable cameras are still shipped to a
CUDA workstation for 3D Gaussian Splatting (3DGS) training. Training on the
device that captured them, or on its companion phone or laptop, requires
more than translating kernels: GPUs and drivers differ in parallel features,
memory limits, and command submission, and these differences decide whether
a training schedule executes and returns the correct update. We present Cross3DGS, a cross-vendor, cross-platform Vulkan trainer that
keeps one optimizer and chooses its execution schedule per device in two
steps: rules on device limits, workload size, and observed driver failures
exclude the schedules known not to run correctly, and a fixed preference
order selects one of the remaining feasible schedules. The same core completes standard-quality
training (30k updates, one million Gaussians) on every configuration we measured, spanning NVIDIA, Intel,
Arm, Qualcomm, AMD, and Apple GPUs and including Android devices and an
iPhone, whereas the existing portable trainers VkSplat and Brush do not run
on some of the mobile GPUs. Averaged over 11 scenes on a workstation,
Cross3DGS has higher quality than VkSplat with no
significant difference in training time, and Brush reports about three times
the training time at comparable quality. The final quality of Cross3DGS is nearly identical
across devices. A mobile application turns a handheld capture
into a viewable model entirely on the device.
\end{abstract}

\section{Introduction}
\label{sec:intro}

3D Gaussian Splatting (3DGS)~\cite{kerbl3dgs} makes radiance-field
reconstruction fast enough that the bottleneck has moved from optimization
time to where optimization can run. Captures from phones, tablets, smart
eyewear, and other wearable cameras are still transferred to a workstation
with a CUDA GPU.
Training on the capture device or its companion phone or laptop would keep
private spaces on the device, remove the round trip, and work offline. We
target this as batch work after a capture, for example while the phone
charges: on our phones training takes minutes to hours.
Commercial and research systems already move parts of this pipeline
off the workstation~\cite{scaniverse,vksplat,brush,pocketgs},
but no trainer we are aware of completes standard-quality training (30k
updates, up to one million Gaussians) on all eight desktop, laptop, phone, and
tablet GPU configurations that we measure, which span six vendors.

Writing the kernels in a cross-vendor shading language does not by itself
make the trainer portable~\cite{donaldson2017shader,decoupledfallback}.
Each step of 3DGS training translates to portable compute kernels, yet the
execution \emph{schedule} around them does not port. Subgroup width, float32 atomic
add, the storage-buffer range, and the reliability
of batched command submission differ across our devices, and they decide
whether a nominally equivalent step completes and returns a correct update.
The Vulkan trainer VkSplat~\cite{vksplat} needs subgroups of at least 32 lanes and has
no mobile host, and the WebGPU trainer Brush~\cite{brush} exceeds the
storage-buffer range of the \gAdreno{} GPU; of the eight configurations we
measure, they complete standard-quality training on four and six
(Figure~\ref{fig:high-low-overview}).

Cross3DGS therefore keeps one optimizer and a catalog of execution
schedules, and chooses a schedule for each device in two steps. Step~1
excludes the schedules known not to run correctly on the device: those that
exceed its limits or the workload, and those that failed on its driver.
Step~2 selects one of the remaining feasible schedules with a fixed
preference order. Offline tests confirm that the gradient
methods in the catalog compute the same update within floating-point
tolerance.
The contributions are:
\begin{itemize}
  \item A cross-vendor, cross-platform Vulkan~1.1 trainer that chooses a schedule per device with 13
  explicit feasibility rules and a fixed preference order, together with
  offline checks that every gradient method computes the same update
  (Secs.~\ref{sec:formulation} and~\ref{sec:lowering}).
  \item Full standard-quality training on all eight measured
  configurations, including two Android devices and an iPhone, with
  final quality agreeing within 0.07~dB across devices on two scenes
  (Sec.~\ref{sec:eval}).
  \item An explicit account of what portability costs: an 11-scene
  workstation comparison with VkSplat and Brush, and the measured cost of a preference order
  that never times schedules on the device (Sec.~\ref{sec:eval:cost}).
  \item An Android and iOS capture-to-view application that runs capture,
  structure from motion, training, and viewing on the device
  (Figure~\ref{fig:teaser}, Sec.~\ref{sec:eval:mobile}).
\end{itemize}

\noindent\textbf{Code release.} The trainer, the Android and iOS
applications, and the benchmark and evaluation scripts are available
at \url{\paperrepourl}.

\section{Related Work}
\label{sec:related}

\paragraph{Portable and on-device 3DGS training.}
The original 3DGS system~\cite{kerbl3dgs} couples differentiable
rasterization with adaptive densification; MCMC
densification~\cite{mcmc3dgs} provides a fixed-capacity alternative.
Trainers such as gsplat~\cite{gsplat}, LichtFeld
Studio~\cite{lichtfeld2025}, and Postshot~\cite{postshot} require CUDA-capable NVIDIA GPUs. OpenSplat~\cite{opensplat}
reaches several vendors through per-backend kernels without a Vulkan or
Android path, and Spirula Studio~\cite{spirula} covers desktop GPUs only.
PocketGS~\cite{pocketgs} trains for 500 updates through Metal on one iPhone
from an on-device dense prior. Mobile-GS~\cite{mobilegs} and Texture3dgs~\cite{texture3dgs}
evaluate on Snapdragon phones but run only rendering there and train on
desktop NVIDIA GPUs. XPR~\cite{xpr} and 3DGS.cpp~\cite{threedgscpp} are
renderers, the first for several accelerators and the second for Vulkan, and
Apple-specific routes train through
Metal, MLX, or MPS~\cite{gaussiansplattingmlx,msplat,gsplatmlx,voxelio,radiancekit,threedsplatapp}.
VkSplat~\cite{vksplat} and Brush~\cite{brush} are the closest portable
trainers (Vulkan and WebGPU) and serve as our baselines. Feed-forward
models~\cite{pixelsplat,mvsplat} skip per-scene training, and
PocketSplat~\cite{pocketsplat} runs one on an iPhone from four posed views,
but they rely on large-scale pretraining and are evaluated on a few input views; we address
per-scene training, which assumes no prior about the scene.

\begin{figure*}[t]
  \centering
  \includegraphics[width=0.92\textwidth]{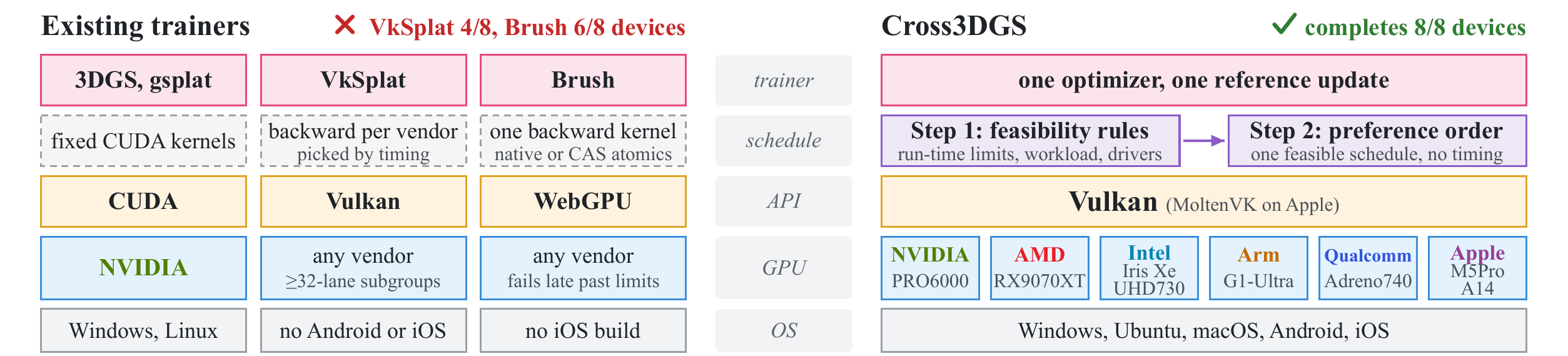}
  \caption{Existing 3DGS trainers and Cross3DGS as software stacks. Existing
  trainers fix their kernel sets per backend or vendor, or pick among them by
  timing; Cross3DGS selects a schedule for each device in two steps and is the
  only one of the three trainers we ran to complete standard-quality
  training on all eight devices (Figure~\ref{fig:high-low-overview}).}
  \label{fig:trainer-stack}
\end{figure*}

\paragraph{How Cross3DGS differs from existing trainers.}
Figure~\ref{fig:trainer-stack} contrasts Cross3DGS with existing 3DGS
trainers. VkSplat compiles its shaders for a subgroup width of 32, so it
rejects a device with narrower subgroups, and it emulates a missing float32
atomic add only after the user edits a macro and recompiles. At startup it
checks device limits such as subgroup width and shared memory, and it drops
the gradient kernels that exceed shared memory before choosing among the
rest by Thompson sampling over measured times. Its checks thus change one
option, the gradient kernel. Cross3DGS gives each of its six options an
alternative value, so a failed check moves the schedule to it, and its rules
read two inputs that device limits do not cover: the workload size and
failures observed on specific drivers. It then selects a schedule without
timing it on the device.

\paragraph{Schedule search and feasibility.}
Compiler systems separate algorithms from schedules
\cite{halide,tiramisu,tvm}, and Ansor~\cite{ansor} and Sassie
et al.~\cite{sassie2026vulkan} search schedules by measuring them on the
target. Cross3DGS runs no search on the target: explicit rules decide
which schedules are feasible, and a fixed order picks one. Differentiable
programming and graphics systems provide or generate derivatives across frameworks, shaders, and
kernels~\cite{nvdiffrast,drjit,slangd,dressi,enzyme,difftaichi}; our
gradients are hand-written kernels.

\paragraph{Gradient accumulation and validity.}
3DGS gradient optimizations reduce scatter contention through warp
aggregation or per-splat ownership~\cite{distwar,taming3dgs}, and
Faster-GS~\cite{fastergs} separates implementation changes from fidelity
trade-offs. We treat these ownership strategies as alternative schedules of
one update. A floating-point sum depends on its
order~\cite{collange2015reduction}, and a driver can mis-execute a feature
it reports as supported~\cite{donaldson2017shader}; a gradient wrong for
either reason can still reach a plausible PSNR, so we test the update itself.

\section{Choosing a Schedule for Each Device}
\label{sec:formulation}

\begin{figure*}[t]
  \centering
  \includegraphics[width=0.94\textwidth]{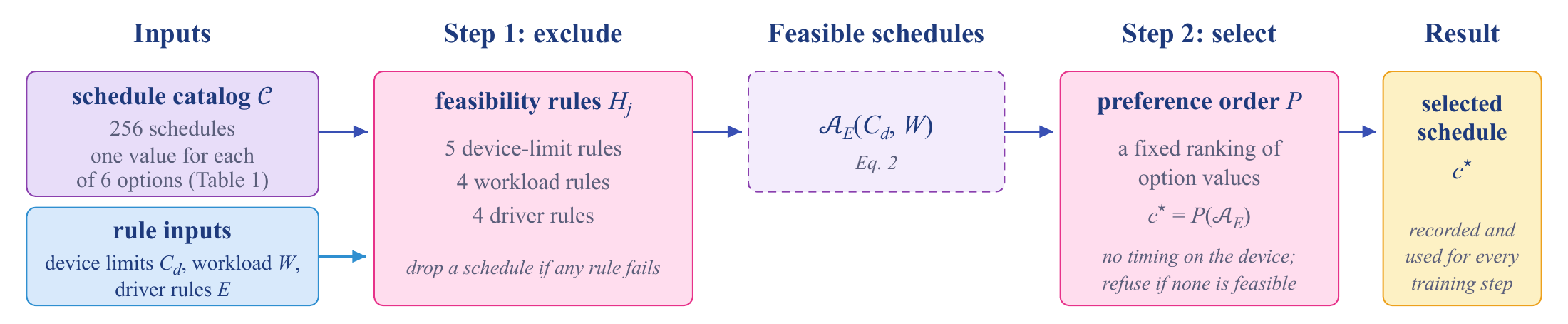}
  \caption{Choosing a schedule for a device. Step~1 drops every schedule in
  the catalog that fails a feasibility rule; the rules read the device
  limits $C_d$, the workload $W$, and the driver rules $E$. Step~2 picks one
  of the remaining feasible schedules with a fixed preference order. The
  gradient methods in the catalog are checked offline against the reference
  update. Table~\ref{tab:schedules} shows both steps on the eight measured
  devices.}
  \label{fig:constraint-lowering}
\end{figure*}

\paragraph{One update, many schedules.}
Let $\theta$ collect the trainable Gaussian parameters and optimizer state.
For a sampled camera and target image $I$, one training step is
\begin{equation}
  \theta^{+}=U(\theta,I),
  \label{eq:reference-update}
\end{equation}
where $U$ comprises projection, tile binning and sorting, alpha compositing,
the loss, its gradient, and Adam. MCMC-based refinement~\cite{mcmc3dgs}, which
adds and relocates Gaussians every 100 updates, lies outside $U$ and is not
covered by the consistency test below. A \emph{schedule} fixes how the GPU
executes $U$ by choosing one value for each of the six \emph{options} in
Table~\ref{tab:options}. All schedules run the same operations; they differ
in parallel work layout, memory placement, and the order in which
floating-point sums are added. The \emph{schedule catalog} $\mathcal{C}$
holds all 256 combinations. Cross3DGS picks a schedule for each
device in two steps (Figure~\ref{fig:constraint-lowering}).

\begin{table*}[t]
  \caption{The six options of a schedule, the number of values of each, and
  the values. The catalog contains every combination. Within each option,
  values are listed in the order the common preference order ranks them. The last
  three columns name what moves a device to a later value; the driver
  failures were observed on Qualcomm \gAdreno{}.}
  \label{tab:options}
  \centering\scriptsize
  \setlength{\tabcolsep}{4pt}
  \begin{tabular}{@{}lr>{\raggedright\arraybackslash}p{0.37\textwidth}%
      >{\raggedright\arraybackslash}p{0.13\textwidth}%
      >{\raggedright\arraybackslash}p{0.13\textwidth}%
      >{\raggedright\arraybackslash}p{0.12\textwidth}@{}}
    \toprule
    & & & \multicolumn{3}{c}{moved to a later value by} \\
    \cmidrule(l){4-6}
    option & \# & values & device limit & workload & driver failure \\
    \midrule
    \begin{tabular}[t]{@{}l@{}}gradient method\\
      (work assignment\\\ + merge)\end{tabular}
      & 8 & tiled + native atomic,
      tiled + CAS, pixel/subgroup + native atomic,
      pixel/subgroup + CAS, per-splat + native atomic, per-splat + CAS,
      pair-partial + fixed reduction (exact pair gather),
      per-splat + fixed reduction (pair reduction)
      & subgroup width, shared memory, float32 atomic add & -- & scatter
      failures \\
    command submission & 2 & batched (8 updates), fenced (per pass)
      & -- & -- & watchdog timeout \\
    training images & 2 & resident (all in GPU memory), streamed (one per
      update)
      & -- & images exceed one storage buffer & device fault \\
    radix sort & 2 & subgroup, portable
      & -- & -- & corrupted sort keys \\
    projection and Adam & 2 & fused, split
      & buffer bindings & SH coefficients exceed one storage buffer & -- \\
    pair dispatch & 2 & indirect, direct
      & \multicolumn{3}{>{\raggedright\arraybackslash}p{0.38\textwidth}@{}}{none
      yet; direct is kept for drivers where indirect fails or is slower} \\
    \midrule
    total & 256 & $8\times2\times2\times2\times2\times2$ schedules
      & & & \\
    \bottomrule
  \end{tabular}
\end{table*}

\paragraph{Step 1: exclude infeasible schedules.}
A \emph{feasibility
rule} $H_j^{E}(c,C_d,W)$ is 1 when schedule $c$ passes rule $j$ and 0
otherwise. The rules read three inputs: the \emph{device limits} $C_d$
reported by the Vulkan driver~\cite{vulkanspec,spirvspec} (subgroup width,
shared memory, storage-buffer range, float32 atomic add, vendor
identity); the \emph{workload} $W$ (Gaussian capacity,
image size, and number of training images); and the \emph{driver rules} $E$,
failures we observed on a specific driver that its reported limits do not
predict. The \emph{feasible schedules} are those that pass every rule:
\begin{equation}
  \mathcal{A}_{E}(C_d,W)=
  \{c\in\mathcal{C}\mid H_j^{E}(c,C_d,W)=1,\ \forall j\}.
  \label{eq:valid-schedules}
\end{equation}

\paragraph{Step 2: select one feasible schedule.}
A fixed \emph{preference order} $P$ ranks the values of each option, for
example the native atomic merge before compare-and-swap, and picks the selected
schedule $c^\star=P(\mathcal{A}_{E}(C_d,W))$. $P$ was set from offline
measurements on the development devices. One order serves every vendor
except AMD and Intel, whose rankings of gradient methods differ; vendors with no
development device use the common order. Cross3DGS runs no timing on the
device; Sec.~\ref{sec:eval:cost} measures what this costs. If no schedule is
feasible, Cross3DGS refuses to train.

\paragraph{Why feasible schedules compute the same update.}
Different schedules add floating-point numbers in different orders, so they
cannot match bit for bit. With $g$ the reference parameter gradients and
$\Delta\theta=U(\theta,I)-\theta$ the reference Adam update, a schedule is
\emph{numerically consistent} when every tested value $k$ satisfies
\begin{equation}
  |\widehat q_{s,k}-q_k|
  \leq \epsilon_{\mathrm{abs}}^{q}
  +\epsilon_{\mathrm{rel}}^{q}|q_k|,
  \qquad q\in\{g,\Delta\theta\},
  \label{eq:semantic-contract}
\end{equation}
where $\widehat q_s$ is the value computed by schedule $s$. Offline, we test
every gradient method that is feasible on a device on that device, and one
schedule per device, normally the selected one, against an independent CPU
implementation
(Sec.~\ref{sec:eval:semantics}). Choices that change what is optimized, such
as the SH degree, the density strategy, or a sampled gradient estimator, are
not options of a schedule~\cite{fastergs,celarek2025accurate}.

\section{Options, Rules, and the Selected Schedules}
\label{sec:lowering}

Cross3DGS runs $U$ and MCMC refinement as Vulkan compute passes for
projection, binning and radix sorting, rasterization, its gradient, Adam, and
refinement. Parameters,
optimizer state, and the active Gaussian count stay on the GPU in
fixed-capacity buffers, so the command buffers are recorded once and replayed
without per-update allocation or CPU readback. All shaders are compiled to
SPIR-V~1.3~\cite{spirvspec} for Vulkan~1.1~\cite{vulkanspec}.

\begin{table*}[t]
  \caption{The eight measured configurations, their device limits $C_d$, the
  number of the 256 schedules still feasible after each kind of feasibility
  rule is applied in turn, and the selected schedule $c^\star$ for the
  standard-quality setting, one row per option of Table~\ref{tab:options}.
  \gAdreno{} also stores its SH coefficients in two buffers. Memory is dedicated GPU memory for discrete GPUs
  and unified system memory otherwise. Host devices: \gIris{} is integrated in a Core i7-1370P and \gUHD{} in a
  Core i3-14100, whose launch years are given; \gGone{} is in a Xiaomi 17T Pro,
  \gAdreno{} in a Xiaomi Pad 6S Pro 12.4, \gMfive{} in a MacBook Pro, and
  \gAfourteen{} in an iPhone 12 Pro.
  Daggers mark configurations held out
  from the development of the rules and the preference order.}
  \label{tab:schedules}
  \centering\scriptsize
  \setlength{\tabcolsep}{1.6pt}
  \newcommand{\hostsz}{\fontsize{6}{7}\selectfont}
  \begin{tabular}{llcccccccc}
    \toprule
    & vendor & NVIDIA & Intel & Intel & Arm & Qualcomm & AMD & Apple$^\dagger$
      & Apple$^\dagger$ \\
    & product
      & \gPRO{}~\cite{nvidiaRtxPro6000}
      & \gIris{}~\cite{intelI71370p}
      & \gUHD{}~\cite{intelI314100}
      & Mali \gGone{}~\cite{armMaliG1Ultra}
      & \gAdreno{}~\cite{qualcommSnapdragon8Gen2}
      & \gRX{}~\cite{amdRx9070Xt}
      & \gMfive{}~\cite{appleMacBookProSpecs}
      & \gAfourteen{}~\cite{appleIPhone12ProSpecs} \\
    & launch year & 2025 & 2023 & 2024 & 2025
      & 2022 & 2025
      & 2026 & 2020 \\
    & host & \hostsz\twol{Windows}{workstation} & \hostsz\twol{Windows}{laptop iGPU}
      & \hostsz\twol{Ubuntu}{desktop iGPU} & \hostsz\twol{Android}{phone}
      & \hostsz\twol{Android}{tablet} & \hostsz\twol{Ubuntu}{desktop}
      & \hostsz\twol{macOS}{laptop} & \hostsz\twol{iOS}{phone} \\[4pt]
    & memory (GB) & 96 & 16 & 16 & 12 & 12 & 16 & 48 & 6 \\
    \midrule
    \multirow{4}{*}{\shortstack[l]{\emph{Device}\\\emph{limits $C_d$}}}
    & subgroup width & 32 & 32 & 32 & 16 & 64 & 64 & 32 & 32 \\
    & max shared memory (KiB) & 48 & 32 & 64 & 32 & 32 & 64 & 32 & 32 \\
    & float32 atomic add & yes & no & no & no & no & yes & yes & yes \\
    & max storage-buffer range & 4 GiB & 1 GiB & 4 GiB & 2 GiB & 128 MiB
      & 4 GiB & 4 GiB & 3 GiB \\
    \midrule
    \multirow{3}{*}{\shortstack[l]{\emph{Feasible}\\\emph{schedules}\\\emph{(of 256)}}}
    & after device-limit rules & 256 & 128 & 160 & 128 & 128 & 256 & 192
      & 192 \\
    & after workload rules & 256 & 128 & 160 & 128 & 32 & 256 & 192 & 192 \\
    & after driver rules & 256 & 128 & 160 & 128 & 6 & 256 & 192 & 192 \\
    \midrule
    \multirow{7}{*}{\shortstack[l]{\emph{Selected}\\\emph{schedule $c^\star$}}}
    & gradient method & \twol{tiled}{+ native atomic}
      & \twol{pixel/subgroup}{+ CAS} & \twol{pixel/subgroup}{+ CAS}
      & \twol{pixel/subgroup}{+ CAS}
      & \twol{per-splat}{+ CAS} & \twol{per-splat}{+ CAS}
      & \twol{pixel/subgroup}{+ native atomic}
      & \twol{pixel/subgroup}{+ native atomic} \\[5pt]
    & command submission & batched & batched & batched & batched & fenced
      & batched & batched & batched \\
    & training images & resident & resident & resident & resident & streamed
      & resident & resident & resident \\
    & radix sort & subgroup & subgroup & subgroup & subgroup & portable
      & subgroup & subgroup & subgroup \\
    & projection and Adam & fused & fused & fused & fused & split & fused
      & fused & fused \\
    & pair dispatch & indirect & indirect & indirect & indirect & indirect
      & indirect & indirect & indirect \\
    \bottomrule
  \end{tabular}
\end{table*}

\paragraph{Gradient methods.}
The gradient method (first row of Table~\ref{tab:options}) combines a
work assignment with a merge. A \emph{pair} is always a tile--Gaussian pair:
one Gaussian and one 16$\times$16-pixel tile that it overlaps. Every
\emph{work assignment} replays each tile's depth-sorted Gaussians back to
front, as the reference backward pass does~\cite{kerbl3dgs}, and forms
partial gradients for each pair, sums of that Gaussian's gradient terms over
pixels of the tile; they differ in which thread handles which pixel and
Gaussian. In \emph{pixel/subgroup}, lanes hold pixels
and a subgroup sums each Gaussian's terms~\cite{distwar}; \emph{pair-partial} lets one
subgroup cover the tile, so the sum is the pair's complete partial. In
\emph{per-splat}, each thread holds one Gaussian~\cite{taming3dgs} and loops over the tile's
pixels in a staggered order that keeps every pixel back to front. In
\emph{tiled}, a Gaussian's exponent at a pixel is an inner product of two
feature vectors~\cite{tcgs}, so 8 Gaussians are evaluated against all 256
pixels at once~\cite{vksplat}. The \emph{merge} adds the partials into each Gaussian's gradient with
native atomics (float32 atomic add~\cite{vulkanspec}), with compare-and-swap (CAS)
emulation~\cite{cudaguide}, or with a \emph{fixed reduction} that stores every
partial and sums them per Gaussian in a fixed order, without
atomics~\cite{stratton2012scaling}. The two methods that use the fixed
reduction have their own names: pair-partial with it is \emph{exact pair
gather} (exact as opposed to a sampled estimate), and per-splat with it is
\emph{pair reduction}.

\paragraph{Execution options.}
The other five options in Table~\ref{tab:options} change how a step is
executed, not what it computes. \emph{Batched} submission~\cite{vksplat} records eight
updates in one command buffer; \emph{fenced} submits each pass
separately and the host waits for it to finish. \emph{Resident} training images are uploaded
once and indexed each update; \emph{streamed} ones are copied one per update into
a reusable buffer. The \emph{subgroup} radix sort~\cite{merrill2011radix,vksplat} ranks keys
and scans digit counts with subgroup operations; the \emph{portable} one uses
shared memory only.
\emph{Fused} projection and Adam run the projection gradient and the Adam
update in one pass~\cite{vksplat}; \emph{split} uses two passes that bind fewer buffers, as
needed when the spherical-harmonic (SH) coefficients are stored in two
buffers. \emph{Indirect} pair dispatch sizes the tile--Gaussian pair passes
from the current pair count on the GPU~\cite{haar2015gpudriven}; \emph{direct} always launches work
for the full pair capacity and reads no dispatch arguments written on the
GPU. No measured configuration needs direct; it stays in the catalog so that
a driver on which indirect dispatch fails or runs slower can be moved to it
by a new driver rule or preference order, without changing the catalog.

\paragraph{The feasibility rules.}
Each of the 13 rules removes the schedules that use a particular option value
under a particular condition. The five \emph{device-limit rules} read $C_d$
only; for example, native atomic merges require float32 atomic add, and
the tiled assignment requires 48~KiB of shared memory and subgroups of at
least 8 lanes, because one of its exchanges is synchronized only within a
subgroup and spans aligned groups of 8 lanes. The four \emph{workload rules} compare $W$ with $C_d$; for example,
training images stay resident only if all of them fit one storage buffer,
and training is refused if even one of two SH buffers would not fit. The four \emph{driver rules} encode failures observed on a specific
driver whose reported limits allow the schedule; all four are keyed on the
Qualcomm vendor identity. Table~\ref{tab:options} lists
which kind of rule moves each option, and Table~\ref{tab:schedules} shows
how many schedules each kind leaves on every configuration.

\paragraph{The preference order.}
$P$ compares feasible schedules option by option in the order of
Table~\ref{tab:options}, preferring the value listed first: the gradient
method first, then command submission, training images, radix sort,
projection and Adam, and pair dispatch. Every run records the selected
schedule and the full list of feasible schedules. The AMD order ranks
per-splat + CAS first, and the Intel order ranks pixel/subgroup + CAS first
after the native-atomic methods. Each was chosen for speed at one million
Gaussians, the count that standard-quality training reaches, on one device and
driver: \gRX{} for AMD and \gUHD{} for Intel, both under Mesa~25.2.8.

\paragraph{Example: \gAdreno{}.}
Of the eight configurations, only \gPRO{}, \gUHD{}, and \gRX{} meet the tiled requirements. On the \gAdreno{} tablet,
the device-limit rules leave 128 schedules (no float32 atomic add, 32~KiB
of shared memory). The training images and the SH coefficients of one million
Gaussians each exceed the 128-MiB storage-buffer range, so the workload rules
rule out resident images and fused projection and Adam (32 left). The driver
rules then leave 6, three gradient methods each with indirect or direct pair
dispatch. Among them, a one-time offline timing screen chose the per-splat +
CAS gradient method, and the order prefers indirect dispatch.

\section{Implementation and Evaluation Protocol}
\label{sec:implementation}

\paragraph{One training core.}
Desktop, Android, and iOS builds share one C++20 training library on a small
Vulkan compute runtime of our own (custom multi-pass programs, an exclusive
scan, headless context creation, and runtime GLSL-to-SPIR-V compilation);
every gradient is a hand-written kernel. The Apple GPUs run the same
core through MoltenVK~\cite{moltenvk}, and \gRX{} runs it on Mesa RADV.

\paragraph{Evaluation protocol.}
We use seven Mip-NeRF~360 scenes~\cite{mipnerf360}, Train and Truck from
Tanks and Temples~\cite{tanksandtemples}, and DrJohnson and Playroom from
Deep Blending~\cite{deepblending}, at the customary resolutions (outdoor
Mip-NeRF~360 scenes downsampled $4\times$, indoor $2\times$, the others as
distributed), with every eighth image held out. One
evaluator scores the saved held-out renders of every system with PSNR,
SSIM~\cite{ssim}, and LPIPS (AlexNet)~\cite{lpips}. The standard-quality
setting runs 30k updates with SH degree 3, MCMC-based refinement~\cite{mcmc3dgs},
a one-million-Gaussian capacity, and the loss
$0.8\,\mathrm{L1}+0.2\,(1-\mathrm{SSIM})$. On \gPRO{}, each system trains all 11
scenes; across the eight configurations, every system that runs there trains
bicycle, and Cross3DGS also trains garden. Each run is repeated three times
and we report medians. VkSplat~\cite{vksplat} and
Brush~\cite{brush} are run from their public code, one fixed version each,
with their own optimizer trajectories. Each system's training time
keeps its own scope, so time ratios are formed only within the matched
\gPRO{} comparison.

\section{Evaluation}
\label{sec:eval}

\subsection{Where Does Full Training Complete?}
\label{sec:eval:status}

\begin{figure*}[t]
  \centering
  \includegraphics[width=0.98\textwidth]{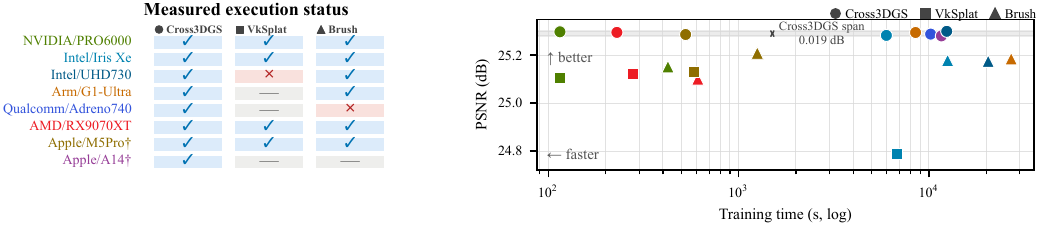}
  \caption{Standard-quality time, quality, and execution coverage on
  \emph{bicycle}. Right: three-run median training seconds (log scale,
  device- and method-local timing scopes) against mean-view PSNR; color
  marks the GPU, as in the text, and marker shape the method. The shaded band
  is the eight-device Cross3DGS PSNR span. Left: check $=$ measured,
  dash $=$ unsupported by the evaluated version (VkSplat: no Android or iOS
  host, $\geq$32-lane subgroups; Brush: no iOS build), cross $=$ invalid
  execution (Brush exceeds the 128-MiB storage-buffer range of \gAdreno{}
  at one million Gaussians; VkSplat hangs the \gUHD{} GPU). Daggers mark the configurations held out from
  the development of the rules and the preference order.}
  \label{fig:high-low-overview}
\end{figure*}

Cross3DGS completes three standard-quality runs on all eight configurations
(Figure~\ref{fig:high-low-overview}, left). VkSplat completes the standard-quality setting on four of the five
desktop and laptop configurations; on \gUHD{} it hangs the GPU within its
first 200 updates (invalid execution). Brush completes it on six. Both baselines are unsupported
or invalid on at least one of the mobile GPUs. On the right of
Figure~\ref{fig:high-low-overview}, Cross3DGS spans about two orders of magnitude in
training time across devices, from 115~s on \gPRO{} to 2.36--3.23~h on the
two Android devices and the iPhone and 3.44~h on the \gUHD{} integrated
GPU, while its final PSNR stays within a 0.019~dB band. Brush reports a longer raw training time than Cross3DGS on
every configuration where both run, including the \gGone{} phone. VkSplat is
within 1\% of Cross3DGS on \gPRO{} and slower on the other three
configurations where both run, with lower PSNR on all four.

\subsection{Do the Selected Schedules Compute the Same Update?}
\label{sec:eval:semantics}

\begin{table}[t]
  \caption{One-step update consistency (Eq.~\ref{eq:semantic-contract}) on
  the six development configurations. \emph{Same GPU}: every feasible
  gradient method against exact pair gather, 100 random small scenes, within
  $10^{-5}+10^{-4}|q_k|$. \emph{CPU double}: the selected schedule against a double-precision implementation
  with finite differences and one Adam update, 20 cases, within
  $10^{-6}+10^{-3}|g_k|$ and $10^{-5}+10^{-2}|\Delta\theta_k|$. Entries are
  the largest absolute differences; viol.\ counts values outside tolerance.}
  \label{tab:consistency}
  \centering\scriptsize
  \setlength{\tabcolsep}{2.4pt}
  \begin{tabular}{lcccccc}
    \toprule
    & feasible & \multicolumn{2}{c}{same GPU} & \multicolumn{2}{c}{CPU double}
      & \\
    \cmidrule(lr){3-4}\cmidrule(lr){5-6}
    configuration & methods & $|\Delta g|$ & $|\Delta\mathrm{Adam}|$
      & $|\Delta g|$ & $|\Delta\mathrm{Adam}|$ & viol. \\
    \midrule
    \gPRO{} & 8 & $3.0\mathrm{e}{-8}$ & $6.3\mathrm{e}{-7}$ & $1.6\mathrm{e}{-5}$ & $3.0\mathrm{e}{-7}$ & 0 \\
\gIris{} & 4 & $9.3\mathrm{e}{-9}$ & $2.4\mathrm{e}{-7}$ & $1.6\mathrm{e}{-5}$ & $1.2\mathrm{e}{-7}$ & 0 \\
\gUHD{} & 5 & $2.8\mathrm{e}{-8}$ & $6.2\mathrm{e}{-7}$ & $1.6\mathrm{e}{-5}$ & $1.2\mathrm{e}{-7}$ & 0 \\
\gGone{} & 4 & $1.4\mathrm{e}{-8}$ & $1.2\mathrm{e}{-7}$ & $1.6\mathrm{e}{-5}$ & $1.2\mathrm{e}{-7}$ & 0 \\
\gAdreno{} & 3 & $9.3\mathrm{e}{-9}$ & $1.2\mathrm{e}{-7}$ & $1.6\mathrm{e}{-5}$ & $1.2\mathrm{e}{-7}$ & 0 \\
\gRX{} & 8 & $3.4\mathrm{e}{-8}$ & $1.5\mathrm{e}{-6}$ & $1.6\mathrm{e}{-5}$ & $1.2\mathrm{e}{-7}$ & 0 \\
\bottomrule

  \end{tabular}
\end{table}

\begin{table}[t]
  \caption{Cross3DGS standard-quality quality on two scenes and eight
  configurations: PSNR/SSIM/LPIPS medians of three runs. Daggers mark held-out configurations.}
  \label{tab:crossdevice}
  \centering\scriptsize
  \setlength{\tabcolsep}{3pt}
  \begin{tabular}{lcc}
    \toprule
    configuration & \emph{bicycle} & \emph{garden} \\
    \midrule
    
\gPRO{} & 25.30/0.767/0.196 & 27.01/0.852/0.108 \\
\gIris{} & 25.28/0.766/0.196 & 26.99/0.852/0.108 \\
\gUHD{} & 25.30/0.767/0.196 & 27.02/0.852/0.108 \\
\gGone{} & 25.30/0.767/0.195 & 27.00/0.852/0.108 \\
\gAdreno{} & 25.29/0.767/0.196 & 26.97/0.852/0.108 \\
\gRX{} & 25.30/0.767/0.196 & 27.03/0.852/0.107 \\
\gMfive{}$^\dagger$ & 25.29/0.767/0.196 & 27.01/0.852/0.108 \\
\gAfourteen{}$^\dagger$ & 25.28/0.767/0.196 & 27.00/0.852/0.107 \\
\midrule
span (dB) & 0.019 & 0.063 \\
\bottomrule

  \end{tabular}
\end{table}

\begin{figure}[t]
  \centering
  \includegraphics[width=\linewidth]{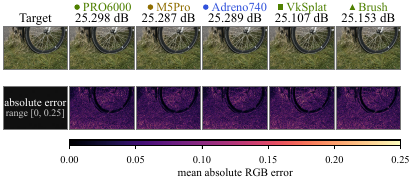}
  \caption{Standard-quality renders and error maps on \emph{bicycle} for
  three devices whose deployed gradient accumulation differs
  (Table~\ref{tab:schedules}) and for the two baselines on \gPRO{}; each
  label reports its run's mean-view PSNR, with the GPU's color and the
  method's marker of Figure~\ref{fig:high-low-overview}. The view is the \gPRO{} per-view
  PSNR median, selected before inspection; one fixed crop is shared by every
  column, and error maps use a fixed $[0,0.25]$ mean-absolute RGB range.}
  \label{fig:qualitative}
\end{figure}

No compared value leaves its tolerance (Table~\ref{tab:consistency}). On
every development configuration, each feasible gradient method matches the
reference method on the same GPU within $3.4\times10^{-8}$ in the gradient and
$1.5\times10^{-6}$ in the Adam update. The CPU comparison checks one schedule per device independently. Its largest
gradient difference is $1.6\times10^{-5}$ on all six GPUs alike, consistent
with error in the CPU's finite differences rather than in any one GPU, and
the GPUs agree with each other far more closely than with it. To confirm that the test can fail, we
make pair reduction on \gPRO{} omit one tile's partial per Gaussian from its
sum: the same 100 cases then report 29,921 gradient violations instead of
none.

Complete runs then measure accumulated behavior. Final PSNR agrees within
0.019~dB on \emph{bicycle} and 0.063~dB on \emph{garden} across all eight
configurations (Table~\ref{tab:crossdevice}), less than the three-run
variation on most single devices (0.013 to 0.097~dB on
\emph{bicycle}). SSIM and LPIPS vary by at most 0.001, and renders from devices with
different gradient accumulation are visually indistinguishable
(Figure~\ref{fig:qualitative}). \gMfive{} and \gAfourteen{} were held out
while the rules and the preference order were developed; both receive the
predicted schedule, pass every CPU-oracle case, and complete every run without any change to the catalog, the rules,
or the preference order.

\begin{table}[t]
  \caption{Standard-quality 11-scene comparison on \gPRO{}: three-run
  medians, P/S/L $=$ PSNR$\uparrow$/SSIM$\uparrow$/LPIPS$\downarrow$,
  s $=$ training-only seconds (Brush: its own integer timing scope). Bold
  marks the best displayed value per row. ``Macro'' averages the three
  per-dataset scene means; its time is the median scene time.}
  \label{tab:datasets}
  \centering\scriptsize
  \setlength{\tabcolsep}{1.6pt}
  \begin{tabular}{@{}lr@{\hspace{2.5pt}}rr@{\hspace{2.5pt}}rr@{\hspace{2.5pt}}r@{}}
\toprule
 & \multicolumn{2}{c}{Cross3DGS} & \multicolumn{2}{c}{VkSplat} & \multicolumn{2}{c}{Brush} \\
\cmidrule(lr){2-3}\cmidrule(lr){4-5}\cmidrule(lr){6-7}
scene & P$\uparrow$/S$\uparrow$/L$\downarrow$ & s & P$\uparrow$/S$\uparrow$/L$\downarrow$ & s & P$\uparrow$/S$\uparrow$/L$\downarrow$ & s \\
\midrule
\multicolumn{7}{l}{\emph{Mip-NeRF 360}}\\
bicycle & \textbf{25.30}/.767/.196 & 115.1 & 25.11/.760/.201 & \textbf{114.5} & 25.15/\textbf{.769}/\textbf{.188} & 425\\
garden & 27.01/.852/.108 & 110.0 & 26.95/.848/.114 & \textbf{106.3} & \textbf{27.21}/\textbf{.855}/\textbf{.098} & 387\\
stump & \textbf{26.84}/\textbf{.786}/\textbf{.163} & \textbf{110.0} & 26.63/.779/.171 & 113.4 & 26.42/.776/.166 & 397\\
bonsai & 32.35/\textbf{.947}/.120 & \textbf{171.8} & 32.56/\textbf{.947}/.119 & 175.2 & \textbf{32.78}/\textbf{.947}/\textbf{.117} & 440\\
counter & 29.40/\textbf{.916}/.140 & 211.2 & 29.34/\textbf{.916}/.141 & \textbf{182.5} & \textbf{29.45}/\textbf{.916}/\textbf{.137} & 471\\
kitchen & 31.55/.930/.094 & 183.3 & 31.41/.929/.096 & \textbf{164.1} & \textbf{31.89}/\textbf{.932}/\textbf{.091} & 426\\
room & 32.20/\textbf{.929}/\textbf{.144} & 169.6 & 32.13/.928/.146 & \textbf{157.0} & \textbf{32.34}/.928/.146 & 440\\
\multicolumn{7}{l}{\emph{Tanks and Temples}}\\
Train & 22.43/.834/\textbf{.138} & \textbf{144.7} & 22.56/.834/.144 & 147.0 & \textbf{22.76}/\textbf{.842}/.143 & 686\\
Truck & \textbf{26.07}/.891/\textbf{.087} & \textbf{136.8} & 25.96/.891/\textbf{.087} & 139.3 & 26.01/\textbf{.893}/.088 & 566\\
\multicolumn{7}{l}{\emph{Deep Blending}}\\
DrJohnson & \textbf{29.79}/.909/.177 & \textbf{113.6} & 29.59/.909/.183 & 117.4 & 29.67/\textbf{.912}/\textbf{.172} & 353\\
Playroom & \textbf{30.40}/\textbf{.915}/.156 & \textbf{112.5} & 30.30/.914/.159 & 114.2 & 30.29/\textbf{.915}/\textbf{.155} & 311\\
\midrule
Macro & 27.86/.883/.139 & \textbf{136.8} & 27.79/.882/.143 & 139.3 & \textbf{27.90}/\textbf{.885}/\textbf{.138} & 426\\
\bottomrule
\end{tabular}

\end{table}

\subsection{What Does the Fixed Preference Order Cost?}
\label{sec:eval:cost}

Step~1 matters first: the \gPRO{} schedule (tiled with native
atomics) is infeasible on six of the seven other configurations because of
their shared memory or missing float32 atomic add, so a schedule tuned on one
device does not port. On the seventh, \gRX{}, it is feasible but slow
(below). The six schedules feasible on all
eight configurations are the conservative ones that \gAdreno{}'s rules leave,
with per-pass fences, one streamed training image per update, the portable
sort, and split projection and Adam.
Forced one at a time on \gPRO{} with 54k Gaussians, exact pair gather, the
portable sort, and streamed images each more than double the step time
(streaming cuts tracked memory from 995 to 312~MB), while splitting
projection and Adam is nearly free.

Step~2 never times schedules on the device; its order was set offline on the
development devices. On the held-out \gMfive{} and \gAfourteen{}, it still
picks the fastest feasible gradient method at the trained
one-million-Gaussian state, so the fixed order costs no speed in that state. Where the common order
did mispredict, on the development devices \gRX{} and \gUHD{}, its choice was
2.94 and 4.17 times slower than the fastest feasible method, which is why
AMD and Intel have their own orders.

\subsection{Is Portability Competitive on a Workstation?}
\label{sec:eval:baselines}

Against VkSplat, Cross3DGS has the higher PSNR on nine of the eleven scenes
and is better on all three metrics on eight before rounding
(Table~\ref{tab:datasets}). A hierarchical bootstrap over scenes and
seed-matched runs, weighting the 11 scenes equally, puts the PSNR gain at 0.074~dB (95\%
interval 0.002 to 0.14~dB), the SSIM gain at 0.0017 (0.0002 to 0.0034), and
the LPIPS reduction at 0.0035 (0.0019 to 0.0052); the time difference of
4.4~s has an interval that includes zero ($-1.2$ to 11.1~s). On the same average,
Brush's PSNR is 0.041~dB higher (interval $-0.10$ to 0.18~dB), while it
reports about three times the training time.

\subsection{What Does Mobile Training Deliver?}
\label{sec:eval:mobile}

\begin{figure}[t]
  \centering
  \includegraphics[width=\linewidth]{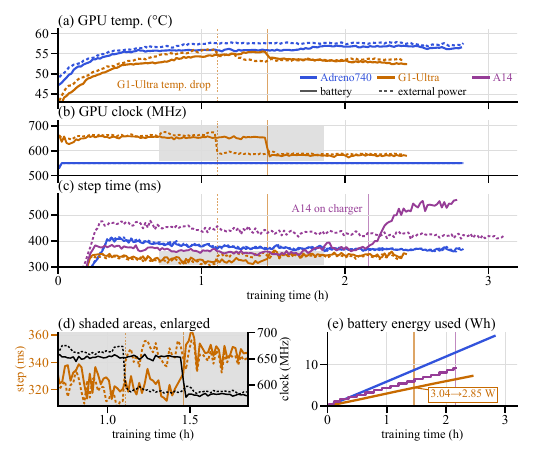}
  \caption{Standard-quality \emph{bicycle} training on the three mobile
  devices, screen on; 1-minute means. (a)~GPU
  temperature, the hottest of eight sensors on \gAdreno{}. (b)~GPU clock.
  (c)~Step time. (d)~\gGone{} in the shaded areas of (b) and (c), enlarged
  and overlaid: step time in orange (left), GPU clock in black (right). (e)~Battery energy since the cable came out;
  \gAfourteen{} from its level; \gGone{} mean power before and after its clock drop. Vertical lines: \gGone{} clock drops,
  \gAfourteen{} on the charger.}
  \label{fig:mobile-power}
\end{figure}

Standard-quality training on the phones and the tablet reaches workstation
quality (Table~\ref{tab:crossdevice}).
Figure~\ref{fig:mobile-power} follows one such run per device on battery and
one on external power, all with the screen on. On
battery it used 16.9~Wh (6.0~W) on \gAdreno{}, 7.2~Wh (3.0~W) on \gGone{}, and
about 9.2~Wh on \gAfourteen{} for 25.5k of the 30k updates. Neither Android
device left the normal thermal status. Heat did not slow \gAdreno{} (GPU
near 57\,\textdegree C at 0.55~GHz). With its GPU at 55 to
56\,\textdegree C, the \gGone{}
clock fell from 0.68 to 0.60~GHz and its mean step time rose 8\%. iOS
reports only a thermal state; \gAfourteen{} was in serious (second hottest
of four) from the fifth minute. Over the first 25.5k updates its mean step
time was 17\% shorter on battery than on external power; its battery then
reached 15\%, and on the charger its median step time rose by 44\%, likely
from charging heat.

\begin{figure}[t]
  \centering
  \includegraphics[width=\linewidth]{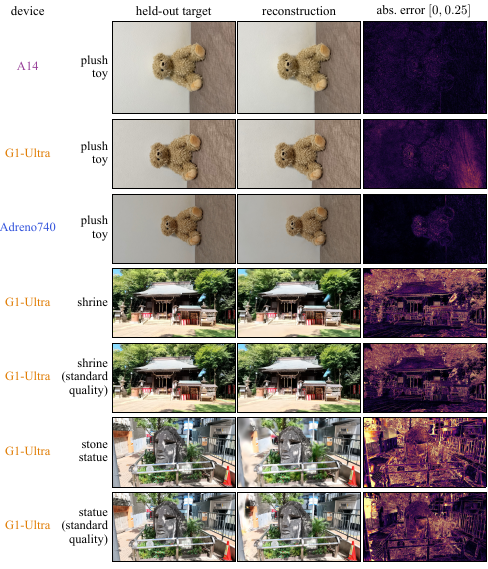}
  \caption{A plush toy reconstructed on \gAfourteen{}, \gGone{}, and
  \gAdreno{}, each from its own capture, and two outdoor scenes on \gGone{},
  at the demonstration preset unless labeled standard quality:
  held-out target, reconstruction, and mean-absolute RGB error in $[0,0.25]$.
  Plush-toy frames are shown as recorded.}
  \label{fig:capture-to-view}
\end{figure}

\begin{table}[t]
  \caption{Each stage of the application for the seven rows of
  Figure~\ref{fig:capture-to-view} (SH3, MCMC); 3DGS times include the live
  preview. Each indoor plush-toy session runs every stage on its device from that
  device's stored capture. The three captures differ, so these
  columns do not compare devices. The outdoor shrine and statue sessions do the
  same from \gGone{} captures; each standard-quality run re-trains from its
  demonstration run's structure from motion and shares its times.}
  \label{tab:capture-to-view}
  \centering\scriptsize
  \setlength{\tabcolsep}{0.7pt}
  \begin{tabular}{lrrrrrrr}
    \toprule
     & \multicolumn{5}{c}{demonstration preset} & \multicolumn{2}{c}{standard quality} \\
\cmidrule(lr){2-6}\cmidrule(lr){7-8}
 & plush toy & plush toy & plush toy & shrine & statue & shrine & statue \\
device & \gAfourteen{} & \gGone{} & \gAdreno{} & \gGone{} & \gGone{} & \gGone{} & \gGone{} \\
\midrule
updates/cap. & 3k/250k & 3k/250k & 3k/250k & 3k/250k & 3k/250k & 30k/1M & 30k/1M \\
registered & 100/100 & 100/100 & 100/100 & 100/100 & 99/100 & 100/100 & 99/100 \\
train/held-out & 87/13 & 87/13 & 87/13 & 87/13 & 86/13 & 87/13 & 86/13 \\
seed points & 62,127 & 95,814 & 66,912 & 54,676 & 36,784 & 54,676 & 36,784 \\
Gaussians & 200,343 & 250,016 & 215,772 & 176,309 & 118,613 & 1M & 1M \\
\midrule
\multicolumn{8}{l}{\emph{time per stage (min)}} \\
extraction & 0.2 & 0.4 & 0.8 & 0.4 & 0.4 & 0.4 & 0.4 \\
SfM features & 4.1 & 1.7 & 1.5 & 1.6 & 1.5 & 1.6 & 1.5 \\
SfM matching & 22.1 & 17.4 & 11.6 & 15.1 & 16.9 & 15.1 & 16.9 \\
SfM mapping & 4.3 & 8.0 & 5.9 & 6.2 & 2.9 & 6.2 & 2.9 \\
3DGS & 15.2 & 7.8 & 17.7 & 6.6 & 6.7 & 264.9 & 219.2 \\
total & 45.9 & 35.3 & 37.5 & 29.9 & 28.4 & 288.2 & 240.9 \\
\midrule
\multicolumn{8}{l}{\emph{held-out quality}} \\
PSNR$\uparrow$ & 31.50 & 28.41 & 31.93 & 18.22 & 15.83 & 19.42 & 16.73 \\
SSIM$\uparrow$ & 0.857 & 0.752 & 0.878 & 0.557 & 0.413 & 0.652 & 0.515 \\
LPIPS$\downarrow$ & 0.277 & 0.293 & 0.324 & 0.460 & 0.721 & 0.225 & 0.365 \\
\bottomrule

  \end{tabular}
\end{table}

The capture-to-view application (Figure~\ref{fig:teaser}) runs every stage on
the device (Table~\ref{tab:capture-to-view}). The Android version records
with ARCore~\cite{arcore}, passes its metric camera positions and intrinsics as
priors to an Android port of CPU COLMAP~\cite{schoenberger2016sfm}, trains
with the same core, and opens a viewer; the iOS version records with
ARKit~\cite{arkit} and shares the rest. At the demonstration preset, a
viewable model is ready 28 to 46 minutes after frame extraction begins, and
structure from motion, not 3DGS, is the longest wait in every session: 19 to
30 minutes against 7 to 18. The \gAfourteen{} session ends in the serious
thermal state. At the standard-quality setting, the two outdoor captures fill
the one-million-Gaussian budget in 3.7 to 4.4 hours of 3DGS on the \gGone{};
both halve the demonstration preset's LPIPS and gain about one decibel
(Figure~\ref{fig:capture-to-view}).

All times in this subsection use each device's default power policy; in
our tests, vendor-specific settings trained up to about twice as fast, with
more heat.

\section{Discussion and Limitations}
\label{sec:limitations}

\paragraph{Fixed preference order.}
Step~2 favors predictable deployment over device-optimal speed. Each
vendor's order was set on one device and driver; others of the same vendor
may rank the methods differently. A one-time timing screen, cached per
device and driver, is the natural extension.

\paragraph{Driver rules.}
All four driver rules are keyed on the Qualcomm vendor identity and were
measured on one \gAdreno{} device and driver. On another Qualcomm GPU
or a newer driver they may exclude schedules that would run, which can cost
speed but not correctness. The held-out
configurations do not test these rules, which needs a second driver. A
driver failure that no rule anticipates is not caught before it occurs, so
the conformance tests are re-run for a new device or driver.

\paragraph{Scope of the evidence.}
Cross-device quality covers two scenes on all eight configurations; the
11-scene comparison is confined to \gPRO{}. Baseline comparisons are
system-level, with different view orders, optimizer state, and
densification trajectories. Less capable devices, such as headsets and
smart eyewear, are left to future work.

\section{Conclusion}

Cross3DGS trains 3DGS with one optimizer on GPUs from six vendors, from
workstations to phones. Feasibility rules choose a schedule each device can
run, numerical tests check its update, and standard-quality training
completes with matching quality on every measured configuration, including
inside a phone application.

\clearpage
{
    \small
    \bibliographystyle{ieeenat_fullname}
    \bibliography{main}
}

\clearpage
\appendix
\raggedbottom
\renewcommand{\topfraction}{0.9}
\renewcommand{\dbltopfraction}{0.9}
\renewcommand{\textfraction}{0.08}
\renewcommand{\floatpagefraction}{0.7}
\renewcommand{\dblfloatpagefraction}{0.7}
\makeatletter
\setlength{\@fptop}{0pt}
\setlength{\@dblfptop}{0pt}
\setlength{\@fpsep}{14pt plus 2pt minus 2pt}
\setlength{\@dblfpsep}{14pt plus 2pt minus 2pt}
\makeatother
\phantomsection
\addcontentsline{toc}{section}{Appendix: Supplementary Material}
\section*{Appendix: Supplementary Material}
This appendix gives implementation and measurement details. Each section
begins by naming the part of the main paper it supplements.

\section{Training settings}
\label{sec:supp-training}
\supplements{the standard-quality setting of Sec.~\ref{sec:implementation}, used for Figure~\ref{fig:high-low-overview} and
Tables~4 and~5, and the demonstration preset of Sec.~\ref{sec:eval:mobile}}

Table~\ref{tab:supp-hyper} defines standard quality; the demonstration
preset changes only the update count and capacity (Sec.~\ref{sec:supp-app}).
The three runs of each standard-quality result use the seeds 20260815,
20260816, and 20260817 for both the initial rotations and the view sampling,
except \emph{bicycle} on \gIris{}, which uses 20260820, 20260821, and 20260822.
The single runs of Figure~\ref{fig:mobile-power} use seed 20260815 on the two Android
devices and seed 42 on the \gAfourteen{}, and the capture-to-view sessions use
the applications' fixed seed 42.

\begin{table}[H]
\caption{Cross3DGS standard-quality settings, identical on every
configuration. $R_i$ and $s_i$ are the rotation and scales of the source
Gaussian.}
\label{tab:supp-hyper}
\centering\scriptsize
\setlength{\tabcolsep}{3pt}
\begin{tabular}{@{}l>{\raggedright\arraybackslash}p{0.61\linewidth}@{}}
\toprule
setting & value \\
\midrule
updates & 30,000, one training image per update \\
view sampling & uniform with replacement, seeded \\
loss & $0.8\,\mathrm{L1}+0.2\,(1-\mathrm{SSIM})$; SSIM with an
  $11\times11$ Gaussian window ($\sigma=1.5$), $C_1=0.01^2$,
  $C_2=0.03^2$, zero padding, mean over the full image \\
SH degree & 3; the active degree is $\min(3,\lfloor t/1000\rfloor)$
  at update $t$ \\
capacity & 1,000,000 Gaussians \\
initialization & SfM points; scale $0.1\times$ the root-mean-square
  distance to the three nearest points; opacity 0.5; random unit
  quaternion; SH DC from the point color, higher SH zero \\
scene scale $s$ & $1.1\times$ the largest camera distance from the
  camera centroid \\
learning rates & position $1.6\times10^{-4}s$, decayed exponentially to
  $1.6\times10^{-6}s$ over 30k updates; scale $5\times10^{-3}$; opacity
  $5\times10^{-2}$; rotation $10^{-3}$; SH DC $2.5\times10^{-3}$;
  higher SH $1.25\times10^{-4}$ \\
Adam & $\beta_1=0.9$, $\beta_2=0.999$, $\epsilon=10^{-15}$, FP32
  moments \\
MCMC-based refinement~\cite{mcmc3dgs} & every 100 updates from 600 to 24,900;
  growth $\times1.05$ per refinement up to the capacity; Gaussians at
  opacity $\le0.005$ are relocated; position noise learning rate
  $5\times10^{5}$; scale and opacity regularizers 0.01 each \\
clone offset & a new clone of Gaussian $i$ moves by
  $0.5R_i\,\mathrm{diag}(s_i)\,z_i$, $z_i\sim\mathcal{N}(0,I_3)$ \\
relocation weight & Eq.~\eqref{eq:supp-weight} \\
rasterization & $16\times16$ tiles; pair buffers for 2,048
  tile--Gaussian pairs per tile on average; opacity clamped at 0.999 \\
\bottomrule
\end{tabular}
\end{table}

The relocation weight of Gaussian $i$ is
\begin{equation}
  w_i=\alpha_i\,\frac{1+1.5\min(\bar g_i/(2\times10^{-4}),\,8)}{1+1.5\cdot8},
  \label{eq:supp-weight}
\end{equation}
where $\alpha_i$ is its opacity and $\bar g_i$ is its mean absolute loss
gradient with respect to the normalized-device-coordinate position since the
previous refinement. The normalization keeps $w_i$ in the range of
opacity-only sampling. Which Gaussians may be relocated still depends on
opacity alone, and the gradient accumulator is cleared at every refinement.
This weight and the clone offset are our changes to MCMC~\cite{mcmc3dgs}, which
samples by opacity alone and adds no offset.

\section{Baselines and measurement}
\label{sec:supp-protocol}
\supplements{the evaluation protocol of Sec.~\ref{sec:implementation} and the comparisons of
Figure~\ref{fig:high-low-overview} and Sec.~\ref{sec:eval:baselines} (Table~\ref{tab:datasets})}

\paragraph{Baselines.}
VkSplat~\cite{vksplat} (\texttt{e26c254}, MCMC) and Brush~\cite{brush}
(\texttt{3b80985}, Vulkan backend) use 30k updates, SH degree 3, up to one
million Gaussians, and SSIM weight 0.2 on the same images, poses, and split
as Cross3DGS. Each retains its own initialization, learning rates,
densification, and optimizer state. VkSplat reshuffles the views every epoch
after the first; its seed changes only this order, while its other random
numbers are the same in every run. Brush uses full input resolution.

\paragraph{Evaluation.}
The saved renders use the split of Sec.~\ref{sec:implementation}. PSNR is the mean of
per-view PSNRs. SSIM~\cite{ssim} uses an
$11\times11$ Gaussian window with a 5-pixel border excluded, and LPIPS uses
the AlexNet backbone~\cite{lpips} on inputs scaled to $[-1,1]$.

\paragraph{Timing and memory scopes.}
Table~\ref{tab:supp-timing} defines each reported training time. The
device-wide allocation delta of Sec.~\ref{sec:supp-memory} is the peak device
memory in use during a run, as reported by \texttt{nvidia-smi}, minus the
value before launch.

\begin{table*}[t]
\caption{Timing scopes. The warm-up affects only step-time percentiles, never the total:
the first 20 updates in the command-line trainer (desktop and iOS), and the first
100 in the Android trainer and the capture-to-view applications.
On Android, the application's training time also includes pipeline upload
and readback.}
\label{tab:supp-timing}
\centering\scriptsize
\setlength{\tabcolsep}{3pt}
\begin{tabular}{p{0.10\textwidth}p{0.22\textwidth}p{0.22\textwidth}p{0.30\textwidth}}
\toprule
method & reported time & included & excluded \\
\midrule
Cross3DGS & sum over training calls of the host wall time
  (\texttt{steady\_clock}) from issuing \texttt{execute} (one update) or
  \texttt{executeBatch} (eight) to its return & all 30,000 updates; the
  host waits for each submission (eight updates when batched, one pass when
  fenced) before issuing the next & scene loading, pipeline and
  shader setup, validation, render and PLY export, evaluation, and viewer
  readback \\
VkSplat & sum over updates of the host wall time (\texttt{perf\_counter})
  around each \texttt{train\_step} call & all
  30,000 updates & the same \\
Brush & the training time its command-line interface reports, in whole
  seconds & what its training loop counts & not controllable from outside \\
\bottomrule
\end{tabular}
\end{table*}

\paragraph{Statistics.}
The differences in Sec.~\ref{sec:eval:baselines} average the three seed-matched
differences within each scene, then average the 11 scenes equally.
Table~\ref{tab:datasets} instead reports scene medians: averaging them gives
0.058 rather than 0.041~dB for Brush minus Cross3DGS in PSNR before rounding.
Table~\ref{tab:supp-seeds} gives per-run PSNR and time. The hierarchical
bootstrap resamples scenes, then seeds within each scene, with replacement:
10,000 resamples, seed 20260802. Brush minus Cross3DGS is 0.0007 in SSIM
($-$0.0022 to 0.0031), $-$0.0019 in LPIPS ($-$0.0043 to 0.0005), and 302~s in
training time (255 to 364~s).

\begin{table}[t]
\caption{Every run of the 11-scene comparison on \gPRO{} (Table~\ref{tab:datasets}). Brush times are its own whole-second scope
(Table~\ref{tab:supp-timing}).}
\label{tab:supp-seeds}
\centering\scriptsize
\setlength{\tabcolsep}{2pt}
\begin{tabular}{llrrrrrr}
\toprule
 & & \multicolumn{3}{c}{PSNR (dB)} & \multicolumn{3}{c}{training time (s)} \\
\cmidrule(lr){3-5}\cmidrule(lr){6-8}
scene & seed & Cross3DGS & VkSplat & Brush & Cross3DGS & VkSplat & Brush \\
\midrule
bicycle & 20260815 & 25.31 & 25.11 & 25.18 & 114.2 & 114.4 & 425 \\
 & 20260816 & 25.30 & 25.11 & 25.15 & 115.3 & 114.8 & 425 \\
 & 20260817 & 25.26 & 25.11 & 25.14 & 115.1 & 114.5 & 426 \\
\addlinespace[1pt]
garden & 20260815 & 27.09 & 26.98 & 27.21 & 110.3 & 106.3 & 386 \\
 & 20260816 & 26.99 & 26.92 & 27.27 & 109.9 & 105.8 & 388 \\
 & 20260817 & 27.01 & 26.95 & 27.18 & 110.0 & 106.4 & 387 \\
\addlinespace[1pt]
stump & 20260815 & 26.84 & 26.63 & 26.42 & 110.0 & 113.4 & 396 \\
 & 20260816 & 26.96 & 26.63 & 26.37 & 109.6 & 113.4 & 397 \\
 & 20260817 & 26.80 & 26.64 & 26.43 & 110.0 & 113.4 & 398 \\
\addlinespace[1pt]
bonsai & 20260815 & 32.30 & 32.40 & 32.81 & 172.4 & 173.7 & 434 \\
 & 20260816 & 32.44 & 32.56 & 32.78 & 171.7 & 175.2 & 442 \\
 & 20260817 & 32.35 & 32.59 & 32.74 & 171.8 & 176.1 & 440 \\
\addlinespace[1pt]
counter & 20260815 & 29.40 & 29.28 & 29.45 & 211.2 & 182.5 & 471 \\
 & 20260816 & 29.38 & 29.37 & 29.45 & 210.7 & 182.5 & 471 \\
 & 20260817 & 29.41 & 29.34 & 29.50 & 212.1 & 183.0 & 471 \\
\addlinespace[1pt]
kitchen & 20260815 & 31.55 & 31.41 & 31.89 & 183.3 & 164.6 & 426 \\
 & 20260816 & 31.66 & 31.60 & 32.03 & 184.9 & 164.1 & 431 \\
 & 20260817 & 31.19 & 31.34 & 31.47 & 183.3 & 162.8 & 426 \\
\addlinespace[1pt]
room & 20260815 & 32.20 & 32.09 & 32.34 & 169.2 & 157.6 & 439 \\
 & 20260816 & 32.17 & 32.13 & 32.25 & 169.6 & 157.0 & 443 \\
 & 20260817 & 32.24 & 32.15 & 32.42 & 170.7 & 156.3 & 440 \\
\addlinespace[1pt]
Train & 20260815 & 22.67 & 22.62 & 22.76 & 143.2 & 146.9 & 689 \\
 & 20260816 & 22.43 & 22.56 & 22.78 & 144.8 & 147.0 & 686 \\
 & 20260817 & 22.38 & 22.42 & 22.37 & 144.7 & 147.9 & 682 \\
\addlinespace[1pt]
Truck & 20260815 & 26.11 & 25.96 & 26.02 & 136.8 & 139.3 & 569 \\
 & 20260816 & 26.07 & 25.95 & 26.01 & 136.9 & 139.6 & 564 \\
 & 20260817 & 26.06 & 26.07 & 25.97 & 136.2 & 138.1 & 566 \\
\addlinespace[1pt]
DrJohnson & 20260815 & 29.47 & 29.59 & 29.73 & 114.3 & 116.7 & 352 \\
 & 20260816 & 29.81 & 29.36 & 29.63 & 113.5 & 118.0 & 353 \\
 & 20260817 & 29.79 & 29.61 & 29.67 & 113.6 & 117.4 & 353 \\
\addlinespace[1pt]
Playroom & 20260815 & 30.39 & 30.47 & 30.22 & 112.2 & 114.7 & 311 \\
 & 20260816 & 30.47 & 30.30 & 30.30 & 112.8 & 113.0 & 310 \\
 & 20260817 & 30.40 & 30.21 & 30.29 & 112.5 & 114.2 & 320 \\
\bottomrule

\end{tabular}
\end{table}

\section{Schedule catalog}
\label{sec:supp-catalog}
\supplements{the six options of Table~\ref{tab:options} and the gradient methods described
in Sec.~\ref{sec:lowering}}

Pairs and gradient methods follow the definitions of Sec.~\ref{sec:lowering}.

\paragraph{Kernel parameters.}
The raster forward uses workgroups of 256 invocations. The gradient passes
use 256 for tiled, one subgroup for pair-partial, and 128 otherwise. Fused
projection and Adam use 32, and the two split passes 256 each. CAS emulation adds a
float32 value through a loop of 32-bit compare-and-swap on its bit pattern.

\paragraph{Contributions.}
A Gaussian enters every tile of its screen-space bounding
rectangle with a 16-bit mask of the 4$\times$4-pixel sub-tiles that its
footprint ($\alpha\ge1/255$) overlaps. At a pixel of a covered sub-tile it
contributes when its exponent is non-positive and
$\alpha=\min(0.999,\,o\,e^{\text{exponent}})\ge1/255$; a pixel stops before
its transmittance would fall below $10^{-4}$, and the forward pass stores how
far into the tile's sorted list the pixel went, its contributor count. Where the clamp at 0.999
is active, only the color receives a gradient.

\paragraph{Work assignments.}
Every work assignment starts from the forward pass's final transmittance and
contributor count of each pixel, replays only the Gaussians within that
count, and recovers the transmittance in front of a Gaussian as
$T/(1-\alpha)$ from the transmittance $T$ behind it. VkSplat~\cite{vksplat} provides per-pixel, per-splat, and
matrix-style variants of this backward pass; the latter two are the origins of
per-splat and tiled. Pixel/subgroup runs 128 lanes per tile with two pixels each;
every subgroup walks the list from the largest contributor count among its
pixels and merges its subgroup sum, so each tile--Gaussian pair receives one
merge per subgroup, the warp-level aggregation of DISTWAR~\cite{distwar}. Per-splat parallelizes over splats as Taming 3DGS
does~\cite{taming3dgs}; it runs 128 lanes per tile and keeps each pixel's
accumulated color and transmittance in shared memory. It takes the tile's
Gaussians from the back in batches of $n\le128$; lane $l$ holds the
$l$-th, and in step $j=0,\dots,n+254$ it updates pixel $j-l$ when
$0\le j-l<256$. A workgroup barrier separates the steps, so every pixel
meets the batch back to front; each pixel's color and transmittance stay in
shared memory across steps and batches. Each lane accumulates its Gaussian's
partial in registers and merges it once, or, for pair reduction, stores it
for the fixed sum described under exact pair gather below. Tiled
writes the logarithm of the unclamped $\alpha$ of Gaussian $i$ with opacity
$o_i$ and conic $Q_i$ at a pixel, $\log o_i-\frac12\mathbf d^\top Q_i\mathbf d$
for the offset $\mathbf d$ from the Gaussian's center, as $\mathbf e(\mathbf u)^\top
\mathbf f_i$: $\mathbf e(\mathbf u)=(1,u_x,u_y,u_x^2,u_xu_y,u_y^2)$ for the
pixel's offset $\mathbf u$ from the tile center, and $\mathbf f_i$ is a
6-vector of the Gaussian, the factorization TC-GS uses to compute alphas as
matrix products~\cite{tcgs}. A workgroup of 256 lanes, one per pixel, takes the
Gaussians 8 at a time from the back; each lane first computes $\alpha T$
and $\partial\mathcal L/\partial\log\alpha$ of its pixel for the 8 in depth
order. The lanes then act as 8 Gaussians $\times$ 32 groups of 8 pixels:
each lane sums its group's terms into the gradient of $\mathbf f_i$ and maps
it to the gradients of the center, conic, and opacity. For each Gaussian,
shared memory then adds its 32 group sums four at a time into 8 partials,
and each of the 8 is merged separately.

\paragraph{Settings that follow from the options.}
Three settings of a run are not options. The higher SH coefficients
(192~bytes per Gaussian) are stored in one buffer, or in two 96-byte shards
when one buffer would exceed the storage-buffer range. In one buffer they are
interleaved in blocks of 32 Gaussians so that neighboring invocations read
contiguous memory. MCMC refinement is dispatched indirectly.

\paragraph{Exact pair gather.}
Let $\mathcal{P}$ be the emitted tile--Gaussian pairs. The gradient of
Gaussian $i$ splits by tile:
\begin{equation}
  \nabla_{\theta_i}\mathcal{L}=\sum_{t:(t,i)\in\mathcal{P}}
  \underbrace{\sum_{x\in t}\frac{\partial \mathcal{L}_x}{\partial\theta_i}}_{p_{ti}}.
  \label{eq:supp-pair-partial}
\end{equation}
Pair-partial gives each lane $256/s$ pixels for subgroup width $s$, so one
subgroup covers a tile; it replays the sorted contributor list as
pixel/subgroup does and writes each $p_{ti}$ to its own pair slot. The slots
reuse the pair emission before sorting: Gaussian $i$ emits $n_i$ pairs, one
per tile of its bounding rectangle, at offsets $b_i$ from an exclusive scan
of the $n_i$, so the slot of tile $t$ is $b_i$ plus the row-major index of
$t$ in the rectangle. Every pair writes its slot, zero where no pixel
contributes, and a second pass, one thread per Gaussian, sums slots $b_i$ to
$b_i+n_i-1$ in that order: the privatization and
scatter-to-gather transformation of general GPU
programming~\cite{stratton2012scaling}. Pair reduction applies the same fixed sum to per-splat
partials. A fixed order removes one source of run-to-run variation but is not
an order-independent summation~\cite{collange2015reduction}. The slots take
three 16-byte vectors per pair, about 400~MB at the pair capacity of
Sec.~\ref{sec:supp-memory}; no selected schedule uses a fixed reduction, so
the tracked total there omits them.

\paragraph{Pair overflow.}
The pair capacity is one buffer size for the whole image, 2,048 pairs per
tile times the number of tiles, not a limit per tile. Pairs beyond it are
not emitted, so their contributions are lost; each update counts them on the
GPU. After training the count of the last update is read back and recorded,
and the evaluation scripts reject a run with a nonzero count.
The Android trainer and capture-to-view application also count, over the
whole run, the updates that overflowed and the largest pair count. Every
Android run and session reported in this paper has
zero overflowing updates; the standard-quality runs of Table~\ref{tab:crossdevice}
and Figure~\ref{fig:mobile-power} peak at 1.80 to 2.84~million pairs, against capacities of 8.31
(\emph{bicycle}) and 8.90~million (\emph{garden}). The desktop and iOS runs
record the count at the last update only.

\section{Feasibility rules}
\label{sec:supp-rules}
\supplements{Step~1 of Figure~\ref{fig:constraint-lowering}, the feasibility rules of Sec.~\ref{sec:lowering}, the last
three columns of Table~\ref{tab:options}, and the device limits and feasible-schedule counts
of Table~\ref{tab:schedules}}

Table~\ref{tab:supp-rules} gives the exact predicates; a missing device
limit fails its rule. Tables~\ref{tab:supp-failures}--\ref{tab:supp-software}
record the driver failures, remaining inputs, and software versions.

\begin{table*}[t]
\caption{The 13 feasibility rules of the catalog. \emph{Kind} is the input a
rule reads: device limits $C_d$, workload $W$, or driver rules $E$;
\emph{option} is the schedule option of Table~\ref{tab:options} that it
constrains. Indirect dispatch is core in Vulkan~1.0, so the
flag is always on and \texttt{DISPATCH-INDIRECT-FEATURE} has never removed a
schedule. Table~\ref{tab:supp-failures} gives the failure behind each driver
rule.}
\label{tab:supp-rules}
\centering\scriptsize
\setlength{\tabcolsep}{3pt}
\begin{tabular}{l l l p{0.34\textwidth} p{0.17\textwidth}}
\toprule
kind & option & rule & a schedule fails when & value instead \\
\midrule
device limit $C_d$ & gradient method & \texttt{SUBGROUP-TILED} & it uses the
  tiled assignment and the subgroup width is below 8 or shared memory is
  below 49,152~bytes & a later gradient method \\
 & gradient method & \texttt{ATOMIC-NATIVE} & it merges with native atomics
  and the device lacks float32 atomic add & CAS or fixed reduction \\
 & projection and Adam & \texttt{BINDING-FUSED} & it fuses projection and
  Adam and a stage may bind fewer than 26 storage buffers & split \\
 & radix sort & \texttt{RADIX-SHARED} & the radix histogram,
  $(\max(2048,256\lceil256/s\rceil)+512)\times4$ bytes for subgroup width
  $s$, exceeds shared memory & none: training is refused \\
 & pair dispatch & \texttt{DISPATCH-INDIRECT-FEATURE} & it dispatches pairs
  indirectly and the runtime's indirect-dispatch flag is off & direct \\
\midrule
workload $W$ & training images & \texttt{DESCRIPTOR-TARGET-RANGE} & it keeps
  images resident and width $\times$ height $\times$ views $\times$ 4 bytes
  exceeds the storage-buffer range & streamed \\
 & projection and Adam & \texttt{BINDING-SHREST-SPLIT} & it fuses projection
  and Adam and capacity $\times$ 192 bytes exceeds the storage-buffer range,
  so the higher SH coefficients are sharded & split \\
 & (SH storage) & \texttt{DESCRIPTOR-SHREST-RANGE} & capacity $\times$ 96
  bytes, one shard, exceeds the storage-buffer range & none: training is
  refused \\
 & training images & \texttt{TARGET-STREAMED-OPT} & it streams images
  outside optimizer training & resident \\
\midrule
driver $E$ & gradient method & \texttt{ATOMIC-SCATTER-QCOM} & on Qualcomm, it
  merges with native atomics or uses tiled or pixel/subgroup with CAS &
  per-splat + CAS or a fixed reduction \\
 & command submission & \texttt{SUBMISSION-BATCHED-QCOM} & on Qualcomm, it
  submits batched & fenced \\
 & training images & \texttt{TARGET-RESIDENT-QCOM} & on Qualcomm, it keeps
  multi-view images resident & streamed \\
 & radix sort & \texttt{RADIX-SUBGROUP-QCOM} & on Qualcomm, it uses the
  subgroup radix sort & portable \\
\bottomrule
\end{tabular}
\end{table*}

\begin{table*}[t]
\caption{Driver failures on the \gAdreno{} tablet (driver 0676.76.1).
Each row ran the option excluded by its rule in Table~\ref{tab:supp-rules}.}
\label{tab:supp-failures}
\centering\scriptsize
\setlength{\tabcolsep}{3pt}
\begin{tabular}{l l p{0.24\textwidth} p{0.36\textwidth}}
\toprule
option & rule & cause: the excluded value that was run & result \\
\midrule
gradient method & \texttt{ATOMIC-SCATTER-QCOM} & a gradient pass merging
  with CAS & scatter failure: the pass timed out and the device was
  lost (\texttt{VK\_ERROR\_DEVICE\_LOST}) \\
command submission & \texttt{SUBMISSION-BATCHED-QCOM} & batched
  submission & watchdog timeout: the run did not finish and was
  stopped by a 480~s watchdog \\
training images & \texttt{TARGET-RESIDENT-QCOM} & all training images
  resident, the current target copied with \texttt{vkCmdCopyBuffer} & device
  fault: the second update faulted, the device was lost, and the tablet
  restarted \\
radix sort & \texttt{RADIX-SUBGROUP-QCOM} & subgroup radix sort & corrupted
  sort keys: of 17,624 sorted entries, 3,727 out of order and 8,580 with wrong
  keys or values \\
\bottomrule
\end{tabular}
\end{table*}

\begin{table}[t]
\caption{Rule inputs not shown in Table~\ref{tab:schedules}, as reported by
each driver. Every configuration allows the 26 bindings that fused
projection and Adam need at the standard-quality setting.}
\label{tab:supp-inputs}
\centering\scriptsize
\setlength{\tabcolsep}{4pt}
\begin{tabular}{lcc}
\toprule
configuration & vendor identity & storage-buffer bindings \\
\midrule
\gPRO{} & \texttt{0x10DE} & 1,048,576 \\
\gIris{} & \texttt{0x8086} & 200 \\
\gUHD{} & \texttt{0x8086} & 201,326,592 \\
\gGone{} & \texttt{0x13B5} & 8,388,608 \\
\gAdreno{} & \texttt{0x5143} & 16,777,216 \\
\gRX{} & \texttt{0x1002} & 8,388,606 \\
\gMfive{} & \texttt{0x106B} & 31 \\
\gAfourteen{} & \texttt{0x106B} & 31 \\
\bottomrule
\end{tabular}
\end{table}

\begin{table}[!t]
\caption{Software of the standard-quality runs (Figure~\ref{fig:high-low-overview}). $^a$Recorded on the same host under the previous Mesa package
(0ubuntu0.24.04.2). $^b$The driver and API were queried on 2026-08-10, before
an over-the-air update to the HyperOS version listed, which the runs used.
$^c$Queried on 2026-11-14 under the same HyperOS version, before the runs.}
\label{tab:supp-software}
\centering\scriptsize
\setlength{\tabcolsep}{3pt}
\begin{tabular}{@{}>{\raggedright\arraybackslash}p{0.16\linewidth}
  >{\raggedright\arraybackslash}p{0.31\linewidth}
  >{\raggedright\arraybackslash}p{0.31\linewidth}
  >{\raggedright\arraybackslash}p{0.12\linewidth}@{}}
\toprule
configuration & operating system & GPU driver & Vulkan API (device) \\
\midrule
\gPRO{} & Windows 11 (10.0.26200) & NVIDIA 596.72 & 1.4.329 \\
\gIris{} & Windows 11 Home (10.0.26200) & Intel 32.0.101.7085 & 1.4.323 \\
\gUHD{} & Ubuntu 24.04.5, Linux 7.0.0-38 & Mesa ANV 25.2.8 (0ubuntu0.24.04.4) & 1.4.318$^a$ \\
\gGone{} & Android 16, HyperOS 3.0.305$^b$ & Arm r54p1$^b$ & 1.3.305$^b$ \\
\gAdreno{} & Android 16, HyperOS 3.0.305 & Qualcomm 0676.76.1$^c$ & 1.3.128 \\
\gRX{} & Ubuntu 24.04.5, Linux 7.0.0-38 & Mesa RADV 25.2.8 (0ubuntu0.24.04.4) & 1.4.318$^a$ \\
\gMfive{} & macOS 26.6.2 (25G83) & MoltenVK 1.4.1, Vulkan SDK 1.4.350.0 & 1.4.334 \\
\gAfourteen{} & iOS 18.7.8 (22H352) & MoltenVK 1.4.1, linked statically & 1.1.334 \\
\bottomrule
\end{tabular}
\end{table}

\paragraph{Workload values.}
On \emph{bicycle}, the 169 packed RGBA8 training images take 687~MB.
At one million Gaussians the higher SH coefficients take 192~MB, or 96~MB
per shard: only the shards fit \gAdreno{}'s 128-MiB storage-buffer range;
all three fit every other configuration.

\section{Preference orders}
\label{sec:supp-orders}
\supplements{Step~2 of Figure~\ref{fig:constraint-lowering}, the preference orders of Sec.~\ref{sec:lowering}, and the
selected schedules of Table~\ref{tab:schedules}; Sec.~\ref{sec:eval:cost} evaluates them}

Table~\ref{tab:supp-orders} gives the gradient-method orders for AMD
(vendor \texttt{0x1002}), Intel (\texttt{0x8086}), and all other vendors
(common). The other options follow Table~\ref{tab:options}.
Table~\ref{tab:supp-regret} reports the held-out timing screens.

\begin{table}[t]
\caption{Gradient-method preference orders. ``Pixel'' is pixel/subgroup
and ``native'' the native atomic merge.}
\label{tab:supp-orders}
\centering\scriptsize
\setlength{\tabcolsep}{3pt}
\begin{tabular}{rlll}
\toprule
rank & common & AMD & Intel \\
\midrule
1 & tiled + native & per-splat + CAS & tiled + native \\
2 & tiled + CAS & exact pair gather & pixel + native \\
3 & pixel + native & per-splat + native & per-splat + native \\
4 & pixel + CAS & pair reduction & pixel + CAS \\
5 & per-splat + native & pixel + CAS & exact pair gather \\
6 & per-splat + CAS & pixel + native & pair reduction \\
7 & exact pair gather & tiled + native & per-splat + CAS \\
8 & pair reduction & tiled + CAS & tiled + CAS \\
\bottomrule
\end{tabular}
\end{table}

\begin{table}[t]
\caption{Held-out timing screens at the trained one-million-Gaussian SH3
state of \emph{bicycle}: 2k updates, three runs per method. Slowdown is the
selected method's step time divided by the fastest feasible method's.}
\label{tab:supp-regret}
\centering\scriptsize
\setlength{\tabcolsep}{3pt}
\begin{tabular}{lllr}
\toprule
configuration & selected method & fastest method & slowdown \\
\midrule
\gMfive{} & pixel/subgroup + native & pixel/subgroup + native & 1.00 \\
\gAfourteen{} & pixel/subgroup + native & pixel/subgroup + native & 1.00 \\
\bottomrule
\end{tabular}
\end{table}

Each vendor order was set by one offline timing screen on one development
device at the trained one-million-Gaussian SH3 state of \emph{bicycle}, with
2k updates and three runs per method, after every method had passed the
same-GPU update-consistency test. AMD ranks its feasible methods by time;
Intel retains the native-atomic methods first in their common relative
order, then ranks the rest by time on \gUHD{}. Only \gRX{} and \gUHD{}
select differently from the common order. On
\gAdreno{}, the same screen of its three feasible gradient methods, each with
the other options of the selected schedule, finds per-splat + CAS fastest;
pair reduction and exact pair gather take 1.04 and 1.42 times as long.

\begin{figure*}[t]
\centering
\includegraphics[width=\linewidth]{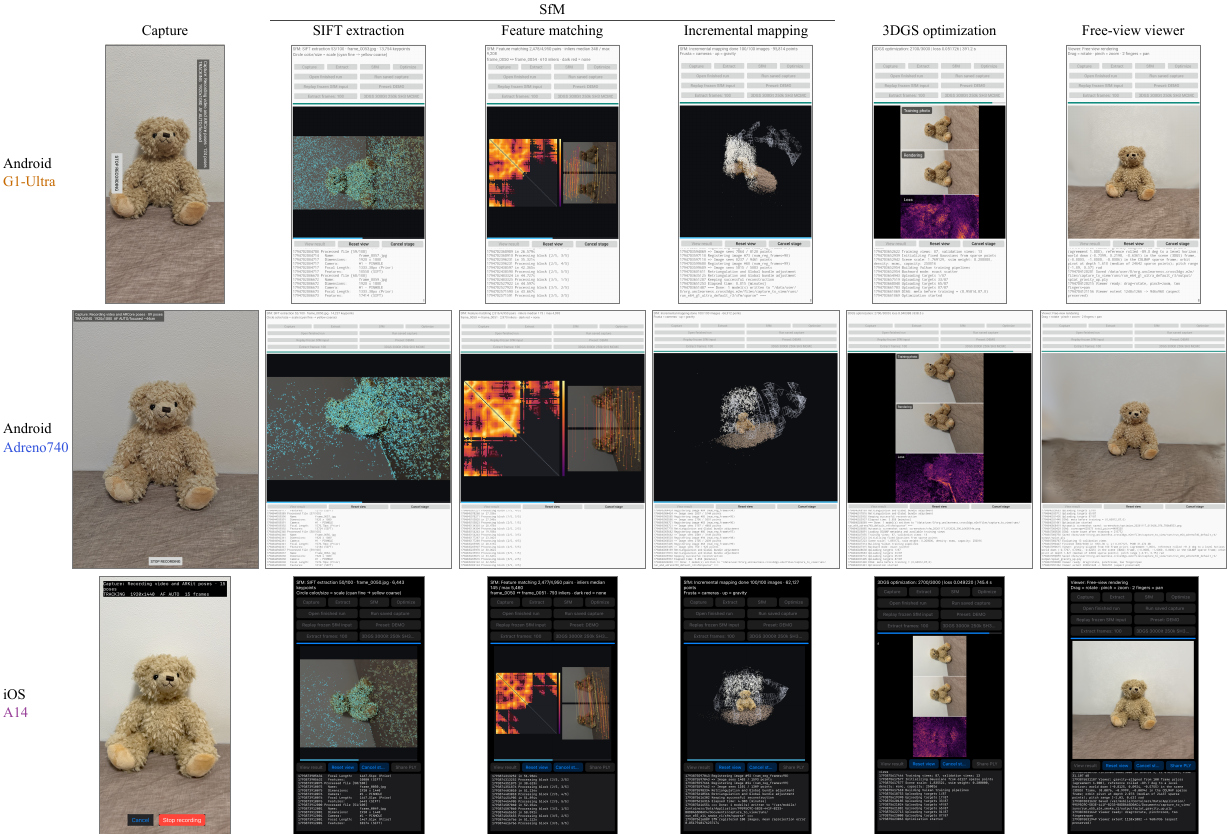}
\caption{Screens from the plush-toy sessions of Table~\ref{tab:capture-to-view}
(100 frames, demonstration preset): \gGone{} (top), \gAdreno{} (middle),
and \gAfourteen{} (bottom). Capture panels were saved during recording;
the other panels during replay. The \gGone{} capture screen is turned
upright. SIFT, matching, and optimization show the frames on their recorded
side; mapping and the viewer orient gravity down.}
\label{fig:supp-app}
\end{figure*}

\section{Update-consistency tests}
\label{sec:supp-consistency}
\supplements{Sec.~\ref{sec:eval:semantics} and Table~\ref{tab:consistency}}

\paragraph{Same-GPU comparison (Table~\ref{tab:consistency}, ``same GPU'').}
Each test case runs one forward pass, loss, gradient, and Adam update on a
$96\times80$ image at SH degree 3. The 100 cases combine four occupancies (4,
12, 32, and 64 Gaussians, from sparse to overlapping) with five cameras and
five parameter states. A state places the Gaussians on a jittered grid
whose extent shrinks and whose Gaussian scale grows with occupancy and draws their depth, scales,
opacity, rotation, color, and higher SH coefficients from a fixed-seed
generator, so every device sees the same cases. Every parameter group is
compared: mean, scale, rotation, opacity, SH DC, and higher SH. Every feasible gradient method is compared with exact
pair gather on the same GPU, and exact pair gather with pixel/subgroup
wherever pixel/subgroup is feasible, which excludes \gAdreno{}.
The tolerances are those of Table~\ref{tab:consistency}.

\paragraph{CPU double-precision oracle (Table~\ref{tab:consistency}, ``CPU double'').}
An independent CPU implementation in double precision evaluates the forward
pass and loss, takes central finite differences with step
$h=\max(10^{-6},10^{-4}|\theta|)$, and applies one Adam update from zero
moments. It uses 20 cases per device, ten at SH degree 0 and ten at SH
degree 3. Each case has two overlapping Gaussians on a $48\times32$ image;
the ten variants shift the camera intrinsics, means, scales, and opacities
so that no finite-difference sample falls on a raster coverage boundary,
and every scalar of every parameter group is differentiated, using the
tolerances of Table~\ref{tab:consistency}.

\paragraph{Two updates with nonzero Adam moments.}
An Adam update from zero moments cannot reveal a gradient wrong only in
scale. A second test compares the second of two updates, taking finite
differences at the state after the first. This can put a finite-difference sample near
a coverage boundary, so when the differences at $h$ and $h/4$ disagree by
more than $10^{-5}+0.05|g|$, the oracle uses $h/16$ and records the scalar
as refined, or as unresolved if $h/4$ and $h/16$ also disagree. On \gPRO{},
all 20 two-update cases pass with the selected schedule; 12 scalars in six
cases were refined and none was unresolved, and the largest Adam-update
difference is $7.9\times10^{-5}$.

\paragraph{Negative control and coverage.}
The negative control of Sec.~\ref{sec:eval:semantics} skips the first pair of every
Gaussian in pair reduction. Besides the reported gradient violations, it
produces 17,146 Adam-update violations and a largest gradient difference of
$1.5\times10^{-2}$; without the change it reports none. Both CPU-oracle tests also run with
the higher SH coefficients forced into two storage buffers, the layout that
the workload rules select when one buffer cannot hold them, and pass on
\gPRO{}. All these tests use the L1 loss alone on a single view: the SSIM
term, the refinement regularizers, training-image residency, and indirect
MCMC dispatch are inactive in them, and complete training runs exercise
them.

\section{Buffer capacities}
\label{sec:supp-memory}
\supplements{the fixed-capacity buffers of Sec.~\ref{sec:lowering} and the baseline
comparison of Sec.~\ref{sec:eval:baselines}}

At the standard-quality setting, Cross3DGS allocates its buffers once for
the full capacity. Each Gaussian slot takes 1,180~bytes for its parameters,
gradients, FP32 Adam moments, forward intermediates, and MCMC bookkeeping,
1.18~GB for one million slots. The sort and pair buffers hold about
8.3~million tile--Gaussian pairs in about 170~MB. On \emph{bicycle}, the
resident images take 687~MB and the per-pixel render, loss, and SSIM buffers
about 90~MB. The total tracked allocation is 2.13~GB. Measured from outside on
\gPRO{}, the device-wide allocation delta of a \emph{bicycle} run
(Sec.~\ref{sec:supp-protocol}) is 2698~MiB for Cross3DGS, against 989~MiB for
VkSplat and 2372~MiB for Brush: Cross3DGS uses the most device memory of the
three, the cost of keeping every buffer resident at full capacity.

\section{Capture-to-view application}
\label{sec:supp-app}
\supplements{the application of Figure~\ref{fig:teaser} and Sec.~\ref{sec:eval:mobile} (Figure~\ref{fig:capture-to-view}, Table~\ref{tab:capture-to-view})}

Figure~\ref{fig:supp-app} shows the application stages of Sec.~\ref{sec:eval:mobile}.

\paragraph{Frame extraction.}
For each target timestamp, both applications decode two uniformly spaced
candidates and keep the one with larger luma-gradient variance. Frames are
saved as JPEG at quality 92, at most 1920 pixels on the long side, with the
AR pose interpolated to the selected frame's timestamp.

\paragraph{Structure from motion.}
CPU COLMAP~\cite{schoenberger2016sfm} uses ARCore~\cite{arcore} or
ARKit~\cite{arkit} camera positions and pinhole intrinsics as priors; the
positions fix metric scale. AR gravity orients the mapping view and viewer;
AR rotations do not initialize mapping.

\paragraph{Training presets.}
The demonstration preset uses the settings of
Table~\ref{tab:supp-hyper} with 3k updates and a capacity of 250,016 Gaussians
(the 250k of Table~\ref{tab:capture-to-view}, rounded up to a multiple of 32), so
its position learning rate decays over the same 30k-update schedule. The
standard-quality setting is Table~\ref{tab:supp-hyper} itself. Every eighth
registered frame is held out for scoring.

\section{Mobile power policy}
\label{sec:supp-policy}
\supplements{the mobile times of Figure~\ref{fig:high-low-overview}, Sec.~\ref{sec:eval:mobile}, and Table~\ref{tab:capture-to-view}}

Table~\ref{tab:supp-power} compares the default Android power policy used
in the main paper with Game Turbo and the same binary under a vendor-listed
benchmark package name (``renamed''), which the policy caps least.
Every Android run and session in this comparison starts at a battery
temperature of at most 33.5\,\textdegree C, without external cooling.

\paragraph{3DGS training only.}
These are \emph{bicycle} runs of 3k updates at SH degree~0 from the
54,275 SfM points, without refinement, two runs per condition. Renaming
halves the \gGone{} step time with PSNR within 0.02~dB, but Game Turbo slows
it down. Both alternatives speed up \gAdreno{} by about 1.25 times. The
higher clocks raise the end-of-training GPU temperature: about
30\,\textdegree C on \gGone{} when renamed and 10\,\textdegree C on
\gAdreno{} under either alternative. iOS exposes no clock; on the \gAfourteen{} an
Instruments trace shows the GPU in its middle performance state for about 72\% of
training, and the step time rose from about 21 to 42~ms as the performance state
fell. The caps likely explain why the \gGone{}, the newest of the three mobile
GPUs, is only 1.2 to 1.4 times faster than the other two in Figure~\ref{fig:high-low-overview}
(2.36~h, against 2.84~h on \gAdreno{} and 3.23~h on \gAfourteen{}).

\paragraph{Capture-to-view application.}
These replay the plush-toy captures of Table~\ref{tab:capture-to-view} at the
demonstration preset, one session per condition; all 100 frames registered.
The default and \gAfourteen{} rows are the sessions of Table~\ref{tab:capture-to-view}. Renaming
speeds up \gGone{} SfM 1.50 times but 3DGS only 1.20 times: its GPU averages
0.69~GHz after SfM, against 1.55~GHz in training alone, possibly from
residual heat. On \gAdreno{}, Game Turbo and renaming speed up 3DGS 1.43
and 1.26 times, both at the same 0.68~GHz. Their smaller SfM gains
come mainly from mapping; matching changes little despite faster CPU clocks.

\begin{table*}[!t]
\caption{Power policies (Sec.~\ref{sec:supp-policy}). Clocks are means
(GHz): CPU during SfM matching, slowest cluster first; GPU during 3DGS. Step is
speed relative to default; stage times are minutes. End temperatures
(\textdegree C) average the last 15~s and both runs for 3DGS only, or the
last 30~s for capture-to-view. GPU temperature is the hottest reported
sensor (one on \gGone{}, eight on \gAdreno{}), so the devices' readings
differ in scope. iOS reports thermal state instead; a dash is unreported.
Spec and AnTuTu give maxima: kernel CPU limits matching vendor figures, and
observed benchmark clocks, respectively. $^a$An Instruments trace places
\gAfourteen{} in the middle of three GPU performance states for about 72\%
of training, the rest at maximum; iOS reports no clock. $^b$Not vendor
specifications: GPU clocks from
Comparigon~\cite{comparigonDimensity9500,comparigonSnapdragon8Gen2};
\gAfourteen{} CPU clocks from The Apple Wiki~\cite{applewikiA14}
(2.99~GHz) and WikiChip~\cite{wikichipA14} (1.82~GHz).}
\label{tab:supp-power}
\centering\footnotesize
\setlength{\tabcolsep}{4pt}
\begin{tabular}{@{}llrrrrcrrrrr@{}}
\toprule
 & & \multicolumn{4}{c}{3DGS only} & \multicolumn{6}{c}{capture-to-view} \\
\cmidrule(lr){3-6}\cmidrule(l){7-12}
 & & & & \multicolumn{2}{c}{temp.} & & & & & \multicolumn{2}{c}{temp.} \\
\cmidrule(lr){5-6}\cmidrule(l){11-12}
device & policy & GPU & step & GPU & battery & CPU & GPU & SfM & 3DGS & GPU & battery \\
\midrule
\gGone{} & spec & & & & & 2.70 / 3.50 / 4.21 & 1.72$^b$ & & & & \\
 & AnTuTu & & & & & 2.70 / 3.50 / 4.21 & 1.72 & & & & \\
 & default & 0.65 & 1 & 45.3 & 32.2 & 1.08 / 1.47 / 1.74 & 0.57 & 27.1 & 7.8 & 53.6 & 40.9 \\
 & Game Turbo & 0.56 & 0.86 & 44.6 & 33.0 & 1.17 / 2.11 / 2.17 & 0.56 & 18.9 & 7.9 & 58.0 & 43.9 \\
 & renamed & 1.55 & 2.0 & 75.7 & 33.0 & 1.64 / 3.03 / 3.36 & 0.69 & 18.1 & 6.5 & 76.2 & 48.9 \\
\addlinespace
\gAdreno{} & spec & & & & & 2.02 / 2.80 / 3.19 & 0.68$^b$ & & & & \\
 & AnTuTu & & & & & 2.02 / 2.80 / 3.19 & 0.68 & & & & \\
 & default & 0.55 & 1 & 50.5 & 32.8 & 1.89 / 1.57 / 1.52 & 0.55 & 19.0 & 17.7 & 51.3 & 33.9 \\
 & Game Turbo & 0.68 & 1.25 & 60.1 & 33.5 & 1.90 / 2.67 / 2.89 & 0.68 & 17.1 & 12.4 & 61.0 & 36.5 \\
 & renamed & 0.68 & 1.26 & 60.4 & 33.8 & 1.90 / 2.67 / 3.08 & 0.68 & 17.0 & 14.1 & 60.9 & 36.2 \\
\addlinespace
\gAfourteen{} & spec & & & & & 1.82 / 2.99$^b$ & -- & & & & \\
 & default & middle$^a$ & 1 & \multicolumn{2}{c}{fair} & -- & -- & 30.5 & 15.2 & \multicolumn{2}{c}{serious} \\
\bottomrule
\end{tabular}
\end{table*}

\begin{table*}[!t]
\caption{Standard-quality \emph{bicycle} runs of Figure~\ref{fig:mobile-power}.
Peaks during training are the highest 1-minute GPU mean and highest battery
reading. Sensors follow Table~\ref{tab:supp-power}. Time covers
training; energy and mean power cover unplug to training end, including the
screen. Neither Android charger kept up; \gGone{} drew battery power even
at 100\%. $^a$Energy over the 2.17~h before charging at 25.5k updates:
battery-level drop times the rated 10.78~Wh, an upper bound for an aged
battery. $^b$Powered from a computer's USB port.}
\label{tab:supp-energy}
\centering\scriptsize
\setlength{\tabcolsep}{4pt}
\begin{tabular}{llcrrrrrrrl}
\toprule
 & & & GPU (\textdegree C) & \multicolumn{2}{c}{battery (\textdegree C)} & & & & & \\
\cmidrule(lr){4-4}\cmidrule(lr){5-6}
device & power & Fig.~\ref{fig:mobile-power} & peak & at launch & peak & time (h) & energy (Wh)
  & mean (W) & battery (\%) & GPU clock \\
\midrule
\gAdreno{} & battery & a, b, c, e & 57.0 & 32.9 & 37.4 & 2.82 & 16.9 & 6.0
  & 100$\to$54 & 0.55~GHz throughout \\
 & external & a, b, c & 57.9 & 32.9 & 36.7 & 2.83 & -- & -- & 91$\to$84
  & 0.55~GHz throughout \\
\addlinespace
\gGone{} & battery & a--e & 55.5 & 29.9 & 41.2 & 2.45 & 7.2 & 3.0
  & 100$\to$77 & 0.60~GHz from 19.8k updates \\
 & external & a--d & 56.1 & 30.3 & 41.3 & 2.44 & -- & -- & 100$\to$100
  & 0.60~GHz from 16.1k updates \\
\addlinespace
\gAfourteen{} & battery & c, e & -- & -- & -- & 2.80 & 9.2$^a$ & 4.2$^a$
  & 100$\to$15 & not reported by iOS \\
 & external$^b$ & c & -- & -- & -- & 3.12 & -- & -- & 100$\to$100
  & not reported by iOS \\
\bottomrule
\end{tabular}
\end{table*}

\section{Temperature and energy}
\label{sec:supp-thermal}
\supplements{the mobile training times of Figure~\ref{fig:high-low-overview}, Sec.~\ref{sec:eval:mobile} with Figure~\ref{fig:mobile-power}, and
the sessions of Table~\ref{tab:capture-to-view}}

\subsection{3DGS training only}
\label{sec:supp-thermal-3dgs}

\paragraph{Thermal state of the reported runs.}
Figure~\ref{fig:high-low-overview} uses external power, screen on at about half brightness,
and no external cooling (computer USB power on \gAfourteen{}). Android
thermal status stayed normal; \gAfourteen{} was mostly serious.

\paragraph{Training on battery.}
Table~\ref{tab:supp-energy} pairs one battery and one external-power run
per device, screen on at about half brightness. Android energy integrates battery
current times voltage, sampled by Perfetto at 1~Hz, from unplug to training
end; iOS~18 reports battery level in 5\% steps. For the \gAfourteen{}
comparison in Sec.~\ref{sec:eval:mobile}, the matched window is the first 25,456
updates: mean step times are 306~ms on battery and 367~ms on external
power. In the battery run, charging starts at 15\% and median step time
rises from 355 to 512~ms.

\paragraph{GPU clock policy.}
The \gGone{} step-time increase in Sec.~\ref{sec:eval:mobile} compares the 5,000
updates before and after its clock drop. Mean battery power is computed
from unplug to the drop and from the drop to training end (3.04 and
2.85~W). Its runs take 3 to 4\% longer than the Figure~\ref{fig:high-low-overview} median;
the \gAdreno{} runs differ from their median by less than 1\%.

\subsection{Capture-to-view application}
\label{sec:supp-thermal-e2e}

Table~\ref{tab:supp-e2e-thermal} gives the temperatures of the three plush-toy
sessions, which ran on external power with the screen on.

\begin{table*}[!t]
\caption{The plush-toy sessions of Table~\ref{tab:capture-to-view}, on external
power. Peak is the highest temperature during the session, reached during SfM
on both Android devices; the GPU temperature is that of
Table~\ref{tab:supp-power}. iOS reports no temperature. The captures differ,
so the rows do not compare devices.}
\label{tab:supp-e2e-thermal}
\centering\scriptsize
\setlength{\tabcolsep}{3.5pt}
\begin{tabular}{@{}lrrrrlr@{}}
\toprule
 & \multicolumn{2}{c}{GPU (\textdegree C)} & \multicolumn{2}{c}{battery (\textdegree C)} & & \\
\cmidrule(lr){2-3}\cmidrule(lr){4-5}
device & at launch & peak & at launch & peak & thermal state & total (min) \\
\midrule
\gAfourteen{} & -- & -- & -- & -- & nominal$\to$serious & 45.9 \\
\gGone{} & 45.2 & 66.5 & 32.4 & 41.8 & normal & 35.3 \\
\gAdreno{} & 42.8 & 70.8 & 32.7 & 37.7 & normal & 37.5 \\
\bottomrule
\end{tabular}
\end{table*}

\FloatBarrier
\section*{Tool-use disclosure}
An LLM-based coding assistant supported implementation, command orchestration,
aggregation, visualization, and copyediting. Reported metrics were computed
from recorded outputs by fixed evaluators and verified by the authors, who
remain responsible for the implementation, experiments, figures, and text.

{\small

\end{document}